\documentclass{article}

\usepackage[dblblindworkshop,final]{neurips_2026}  
\workshoptitle{LIGHT: Lightweight Trustworthy Foundation Models}

\usepackage[utf8]{inputenc}
\usepackage[T1]{fontenc}
\usepackage[hidelinks]{hyperref}
\usepackage{url}
\usepackage{booktabs}
\usepackage{amsfonts}
\usepackage{amsmath}
\usepackage{amssymb}
\usepackage{nicefrac}
\usepackage{microtype}
\usepackage{xcolor}
\usepackage{graphicx}
\usepackage{multirow}
\usepackage[scaled=.92]{helvet}  
\usepackage{sansmath}            
\usepackage{tikz}
\usetikzlibrary{positioning,arrows.meta,calc,fit,shapes.geometric}
\usetikzlibrary{decorations.pathreplacing,shadows.blur}  
\usepackage{xspace}
\usepackage{enumitem}

\newcommand{\RM}{\mathrm{RM}}
\newcommand{\BI}{\mathrm{BI}}
\newcommand{\zRM}{z^{\RM}}
\newcommand{\zBI}{z^{\BI}}
\newcommand{\XMerge}{\textsc{XMerge}\xspace}

\title{XMerge: Cross-Axis Selection and Reconstructive Layer Merging for LLM Depth Compression}

\makeatletter
\newif\ifshowcredits
\if@anonymous
  \newcommand{\authornote}{}%
  \showcreditsfalse
\else
  \newcommand{\authornote}{\thanks{Corresponding author (\texttt{jundongh@alumni.upenn.edu}). Jundong Hu led and carried out the research end to end.}}%
  \showcreditstrue
\fi
\makeatother
\author{%
  Jundong Hu\authornote \\
  PayPal AI \\
  \texttt{jundhu@paypal.com} \\
  \And
  Shekar Ramachandran \\
  PayPal AI \\
  \texttt{sheramachandran@paypal.com} \\
}

\begin{document}
\ifdefined\linenumbers\linenumbers\fi
\maketitle

\begin{abstract}
Removing complete transformer layers preserves a standard serving architecture,
but existing depth-compression methods can lose substantial quality, and the loss
varies unpredictably across models. We introduce \XMerge, a post-training method with two components. Cross-axis
selection identifies a block with low relative-magnitude and angular hidden-state
change, and local boundary reconstruction re-fits the adjacent surviving block to
match the original two-block output. \XMerge uses no task labels or end-to-end
fine-tuning, and it introduces neither architectural changes nor additional
inference-time parameters. Across
seven Llama and Qwen backbones (0.5B--8B), five published baselines, and three
layer-reduction levels, its advantage over baselines is largest at the most
aggressive removal: at $k{=}4$ it ranks first on six of seven backbones on CORE
(a 22-task aggregate) and, separately, on six of seven on MMLU (five of seven on
both at once), while avoiding the large perplexity increases of several competing
operators. In a task-level bootstrap, the 95\% confidence intervals for the three largest
CORE margins exclude zero; the remaining margins are consistent with ties. Across the 14 (model, regime)
cells it is also the only \emph{evaluated} operator that never collapses, ranking
top-2 in both zero-shot and in-context regimes; on a first calibration probe (one
backbone) it is the best-calibrated operator. Ablations show that local reconstruction provides
most of the gain, while cross-axis fusion helps when the two selection axes
disagree. The additional construction cost is recovered through per-token decode
savings after roughly tens of thousands of requests.
\end{abstract}

\section{Introduction}\label{sec:intro}

\paragraph{Motivation.} Transformer depth drives inference latency and memory traffic: decode is
sequential in layers, so removing whole blocks cuts wall-clock latency and key-value (KV) cache traffic. Unlike
unstructured sparsity, which needs sparsity-aware kernels, or width pruning, which changes tensor
dimensions~\cite{frantar2023sparsegpt,sun2024wanda,ma2023llmpruner}, \emph{depth} compression removes
entire layers while leaving hidden size, attention, vocabulary, and the serving interface untouched, so a
shallower standard transformer runs on stock infrastructure. But layer dropping can cause disproportionate
degradation, and prior methods disagree on which layers to compress and how the removed layer is absorbed:
by dropping, averaging, collapsing, or reconstructing.

\paragraph{Depth compression is select-then-merge.} Any depth-compression method decomposes into a
\emph{selector} $\mathcal{S}$ (which block to remove) and a structural \emph{operator} $\mathcal{M}$ (how an
adjacent block absorbs it), $\mathcal{C} = \mathcal{M}\circ\mathcal{S}$. This decomposition separates two
failure modes: selecting an important block and failing to preserve the combined mapping of a safe block and
its neighbor. We evaluate these two sources of error separately.

\paragraph{Proposed solution.} \XMerge implements both decisions explicitly. \emph{Cross-axis
selection} marks a block as safe to remove only when it is quiet on \emph{both} a relative-magnitude axis and
an angular axis. \emph{Reconstructive merging} then merges that block with an adjacent neighbor, replacing the
pair with one standard transformer block whose existing parameters are optimized to reproduce the pair's
original output boundary states. The two components address different sources of degradation: reconstruction preserves the selected pair's
local mapping, while cross-axis selection reduces the chance of selecting a harmful location.

\paragraph{Contributions.}
\begin{itemize}[leftmargin=1.35em,labelsep=0.45em,topsep=2pt,itemsep=1.5pt,parsep=0pt,partopsep=0pt]
  \item \textbf{A reconstructive merge operator that preserves the serving interface.} We collapse an
    adjacent block pair by re-fitting \emph{one existing surviving standard block} (all of its parameters)
    to reproduce the pair's original output mapping, so the compressed model stays an ordinary $L{-}k$-layer
    transformer with no new modules or added inference-time parameters. Among the operators compared here it is
    the only one that reconstructs by re-fitting a \emph{full existing standard (nonlinear) block}: the closest
    module-free neighbor, ReplaceMe~\cite{shopkhoev2025replaceme}, folds a \emph{linear} map into a surviving
    weight, learned-replacement pruning (LLM-Streamline~\cite{chen2025streamlining}) adds a new module, and the
    analytic folds (LaCo/MKA/SWM/CoMe) use no activation fit at all (Table~\ref{tab:operators},
    \S\ref{sec:operator}).
  \item \textbf{A parameter-free cross-axis selector and evaluation.} We combine a relative-magnitude
    and an angular axis with a maximum ($\mathcal{C}=\mathcal{M}\circ\mathcal{S}$, \S\ref{sec:formulation}) and
    evaluate across seven Llama and Qwen backbones, five baselines, and three compression levels, with
    component, regime, recovery, cost, and latency analyses. \XMerge grows increasingly robust relative to
    strong baselines as more layers are removed, at matched inference cost, and is the only evaluated operator
    with zero regime collapses across all 14 (model, regime) cells (\S\ref{sec:results}--\S\ref{sec:limits}).
\end{itemize}

\section{Related Work}\label{sec:related}

\paragraph{Depth compression.} Layer-dropping methods such as ShortGPT~\cite{men2024shortgpt} and
block-pruning approaches~\cite{kim2024shortenedllama,song2024sleb,blockpruner2024} remove transformer blocks
by activation- or importance-based criteria; magnitude-based analyses such as
Prune\&Comp~\cite{chen2026prunecomp} study the hidden-state magnitude gap that layer removal induces, so a
relative-magnitude selection signal has precedent in this line of work. LaCo~\cite{yang2024laco},
MKA~\cite{mka2024}, and SWM~\cite{swm2024} analytically collapse or combine neighboring parameters, while
CoMe~\cite{come2024} and related methods add feature alignment, distillation, or trained replacement modules.
The most closely related operator, ReplaceMe~\cite{shopkhoev2025replaceme}, is calibration-based and
module-free but \emph{linearizes} the removed blocks (a linear map folded into a surviving weight); \XMerge instead re-fits a
full nonlinear surviving block to the two-block boundary mapping (contrasted operator-by-operator below). We use
``layer merging'' for adjacent-layer depth compression, not model merging across separately trained
networks~\cite{wortsman2022modelsoups,ilharco2023taskarithmetic}.

\paragraph{Positioning.} \XMerge uses a short, layer-local post-training \emph{calibration} (self-supervised MSE to the model's own
pre-merge activations) rather than instantaneous structural pruning or end-to-end recovery. The calibration
yields an otherwise standard shallower transformer, similar to post-training
quantization~\cite{frantar2022gptq,lin2024awq}; we therefore describe it as post-training reconstruction
rather than training-free compression.
Local reconstruction against a dense model's own activations is an established post-pruning
paradigm~\cite{wagner2025freelunch,vanderauderaa2024llmsurgeon}; \S\ref{sec:operator} details our
serving-preserving instantiation.

\paragraph{How \XMerge differs from prior operators.} Table~\ref{tab:operators} compares \XMerge
with the five evaluated baselines and with two related methods that we do not run as baselines: the
closest module-free neighbor ReplaceMe~\cite{shopkhoev2025replaceme} and the closest learned-reconstruction
neighbor LLM-Streamline~\cite{chen2025streamlining}. \XMerge is the only operator that reconstructs by
re-fitting a \emph{full existing standard (nonlinear) block} to the adjacent pair's output while keeping the
served model a module-free $L{-}k$ transformer; the analytic folds combine layers in closed form, ReplaceMe
folds a \emph{linear} map into a surviving weight, and LLM-Streamline fits a \emph{new} module. ``Gradient fit''
refers to the operator itself; CoMe's distillation and SWM's low-rank adaptation (LoRA) are separate recovery stages (Setting~B,
\S\ref{sec:setup}), not part of the operator.

\begin{table}[htbp]
  \centering\small
  \caption{Operator design properties (not accuracy). Above the rule: the five evaluated baselines and \XMerge;
  below: ReplaceMe and LLM-Streamline, cited for positioning only (not run). All except LLM-Streamline yield a
  plain $L{-}k$ transformer with zero new inference parameters, so they differ in \emph{how} the removed block
  is absorbed. ``Gradient fit'' is an operator property; CoMe's distillation and SWM's LoRA are separate Setting-B
  stages (\S\ref{sec:recovery}).}
  \label{tab:operators}
  \begin{tabular}{lcccc}
    \toprule
    & \shortstack{Merge into\\surv.\ block} & \shortstack{Gradient\\fit} & \shortstack{Pair-output\\target} & \shortstack{Full-block\\refit} \\
    \midrule
    ShortGPT (drop)                       & $\times$     & $\times$     & $\times$     & $\times$ \\
    LaCo (analytic fold)                  & $\checkmark$ & $\times$     & $\times$     & $\times$ \\
    MKA (analytic merge)                  & $\checkmark$ & $\times$     & $\times$     & $\times$ \\
    SWM (window average)                  & $\checkmark$ & $\times$     & $\times$     & $\times$ \\
    CoMe (head-group concat)              & $\checkmark$ & $\times$     & $\times$     & $\times$ \\
    \textbf{\XMerge (ours)}               & $\checkmark$ & $\checkmark$ & $\checkmark$ & $\checkmark$ \\
    \midrule
    \multicolumn{5}{l}{\emph{Neighbors cited for positioning (not run as baselines):}} \\
    ReplaceMe (block lineariz.)           & $\checkmark$ & $\times$     & $\checkmark$ & $\times$ \\
    LLM-Streamline (learned repl.)        & $\times$     & $\checkmark$ & $\checkmark$ & $\times$ \\
    \bottomrule
  \end{tabular}
\end{table}

\section{Method: XMerge}\label{sec:method}

\begin{figure*}[t]
  \centering
  \definecolor{ink}{HTML}{161A33}
  \definecolor{muted}{HTML}{6B7390}
  \definecolor{line}{HTML}{DDE2EE}
  \definecolor{slate}{HTML}{5B6482}
  \definecolor{rmc}{HTML}{2D6BE4}   
  \definecolor{bic}{HTML}{E8503A}   
  \definecolor{flc}{HTML}{6C56D9}   
  \definecolor{fuc}{HTML}{F2A12B}   
  \definecolor{ftc}{HTML}{14A37A}   
  \definecolor{gry}{HTML}{EEF1F7}
  \definecolor{grybd}{HTML}{C7CEDD}
  \tikzset{
    base/.style={font=\sffamily\sansmath\footnotesize, text=ink},
    card/.style={rounded corners=9pt, fill=white, draw=line, line width=.7pt,
      blur shadow={shadow xshift=0pt, shadow yshift=-2pt, shadow blur radius=3pt, shadow opacity=14}},
    blk/.style={rounded corners=4.5pt, minimum width=2.0cm, minimum height=.6cm,
      inner sep=0pt, line width=1pt, base},
    gblk/.style={blk, fill=gry, draw=grybd, text=muted},
    pill/.style={rounded corners=5pt, inner sep=2.2pt, font=\sffamily\sansmath\scriptsize\bfseries, text=white},
    tag/.style={rounded corners=4pt, fill=white, draw=grybd, line width=.6pt, inner sep=3pt, base},
    grp/.style={rounded corners=7pt, line width=.8pt, dash pattern=on 3pt off 2pt},
    flow/.style={-{Stealth[length=5.5pt,width=5pt]}, line width=1pt, draw=slate},
    big/.style={-{Stealth[length=9pt,width=8pt]}, line width=2.2pt, draw=ink!70},
    note/.style={font=\sffamily\sansmath\scriptsize, text=muted},
    ptitle/.style={font=\sffamily\sansmath\bfseries\small, text=ink, anchor=west},
    kick/.style={font=\sffamily\scriptsize\bfseries, anchor=west},
    badge/.style={circle, fill=ink, text=white, inner sep=0pt, minimum size=.52cm,
      font=\sffamily\bfseries\small},
  }
  \resizebox{\linewidth}{!}{%
  \begin{tikzpicture}[base]
  \def\H{8.3}

  \begin{scope}
  \node[card, anchor=south west, minimum width=6.4cm, minimum height=\H cm] at (0,0) {};
  \begin{scope}[yshift=.6cm]
  \node[badge] at (.55,7.25) {1};
  \node[ptitle] at (.95,7.25) {\textcolor{flc}{Select} -- Score Every Block};
  \node[note, anchor=west] at (.95,6.8) {per-block change $h_\ell\!\to\!h_{\ell+1}$, z-scored};

  \node[note, anchor=west] at (.75,6.24) {block};
  \fill[rmc, rounded corners=1pt] (3.1,6.27) rectangle (3.28,6.45);
  \node[note, anchor=west, text=ink] at (3.34,6.36) {RM (magnitude)};
  \fill[bic, rounded corners=1pt] (3.1,6.03) rectangle (3.28,6.21);
  \node[note, anchor=west, text=ink] at (3.34,6.12) {BI (angular)};
  \node[note, anchor=west, text=ink] at (5.05,6.24) {fused $s$};

  \draw[grp, draw=flc!55, fill=flc!7] (.5,2.82) rectangle (6.2,4.39);
  \draw[grp, draw=bic!60, fill=bic!6]  (.5,2.07) rectangle (6.2,2.79);

  \foreach \i/\nm/\rmv/\biv/\sv in {
    0/{f_{\ell+3}}/.82/.90/.90,
    1/{f_{\ell+2}}/.55/.35/.55,
    2/{f_{\ell+1}}/.22/.28/.28,
    3/{f_{\ell}}/.12/.15/.15,
    4/{f_{\ell-1}}/.25/.85/.85,
    5/{f_{\ell-2}}/.30/.35/.35}
  {
    \pgfmathsetmacro{\y}{5.55-0.78*\i}
    \node[gblk] at (1.75,\y) {$\nm$};
    \fill[gry, rounded corners=1.2pt] (3.1,\y+.03) rectangle (4.7,\y+.21);
    \fill[rmc, rounded corners=1.2pt] (3.1,\y+.03) rectangle ({3.1+1.6*\rmv},\y+.21);
    \fill[gry, rounded corners=1.2pt] (3.1,\y-.21) rectangle (4.7,\y-.03);
    \fill[bic, rounded corners=1.2pt] (3.1,\y-.21) rectangle ({3.1+1.6*\biv},\y-.03);
    \fill[gry, rounded corners=2pt] (5.0,\y-.1) rectangle (5.9,\y+.1);
    \fill[slate!55, rounded corners=2pt] (5.0,\y-.1) rectangle ({5.0+0.9*\sv},\y+.1);
  }
  \fill[ftc, rounded corners=2pt] (5.0,3.11) rectangle ({5.0+0.9*.15},3.31);
  \node[font=\sffamily\sansmath\tiny\bfseries, text=ftc, anchor=west] at (5.2,3.21) {min $s$};
  \fill[bic!80, rounded corners=2pt] (5.0,2.33) rectangle ({5.0+0.9*.85},2.53);

  \node[blk, fill=fuc!24, draw=fuc, text=fuc!55!black] at (1.75,3.99) {$f_{\ell+1}$};
  \node[blk, fill=flc!18, draw=flc, text=flc!85!black] at (1.75,3.21) {$f_{\ell}$};

  \node[pill, fill=flc, anchor=center] at (5.35,4.39) {selected pair};
  \node[pill, fill=bic, anchor=center] at (5.1,2.80) {high angle: rejected};

  \end{scope}
  \draw[line, line width=.6pt] (.5,1.5) -- (5.9,1.5);
  \node[font=\sffamily\scriptsize\bfseries, text=flc, anchor=west] at (.5,1.3) {SELECTION RULE};
  \node[base] at (3.2,.8) {$\displaystyle \ell^{\star}=\arg\min_j\ \max\!\big(z_j^{\mathrm{RM}},\,z_j^{\mathrm{BI}}\big)$};
  \node[note] at (3.2,.26) {quiet on both axes \ $\cdot$ \ no tunable weights};
  \end{scope}

  \begin{scope}[xshift=7.2cm]
  \node[card, anchor=south west, minimum width=6.6cm, minimum height=\H cm] at (0,0) {};
  \begin{scope}[yshift=.6cm]
  \node[badge] at (.55,7.25) {2};
  \node[ptitle] at (.95,7.25) {\textcolor{ftc}{Merge} -- Reconstruct The Pair};
  \node[note, anchor=west] at (.95,6.8) {layer-local activation matching};

  \draw[grp, draw=slate!70, fill=gry!60] (.45,2.62) rectangle (2.95,4.85);
  \node[pill, fill=slate, anchor=west] at (.58,4.85) {frozen};
  \node[blk, fill=flc!18, draw=flc, text=flc!85!black] (t0) at (1.7,3.05) {$f_{\ell}$};
  \node[blk, fill=fuc!24, draw=fuc, text=fuc!55!black] (t1) at (1.7,4.25) {$f_{\ell+1}$};
  \draw[flow] (t0.north) -- (t1.south) node[midway, right=1pt, note] {$h_{\ell+1}$};

  \draw[grp, draw=ftc!80, fill=ftc!7] (3.65,2.95) rectangle (6.15,4.45);
  \node[pill, fill=ftc, anchor=west] at (3.78,4.45) {merged};
  \node[blk, fill=ftc!20, draw=ftc, text=ftc!55!black] (st) at (4.9,3.62) {$\tilde f_{\ell}$};

  \node[tag] (hin) at (3.3,1.85) {$h_{\ell}$};
  \draw[flow] (hin.north) to[out=100,in=-90] (t0.south);
  \draw[flow] (hin.north) to[out=80,in=-90] (st.south);
  \node[tag] (ht) at (1.7,5.45) {$h_{\ell+2}$};
  \node[tag] (hs) at (4.9,5.45) {$\tilde h_{\ell+2}$};
  \draw[flow] (t1.north) -- (ht.south);
  \draw[flow] (st.north) -- (hs.south);
  \draw[<->, dash pattern=on 3pt off 2pt, line width=1pt, draw=bic, >={Stealth[length=4pt,width=4pt]}]
     (ht.east) -- (hs.west) node[midway, above=1pt, font=\sffamily\sansmath\scriptsize\bfseries, text=bic] {MSE};
  \node[note, anchor=south] at (ht.north) {target};
  \node[note, anchor=south] at (hs.north) {prediction};

  \node[draw=slate!60, rounded corners=5pt, inner sep=2.6pt, font=\sffamily\sansmath\scriptsize\bfseries, text=slate, anchor=south east, line width=.7pt]
    at (6.2,1.7) {$\circlearrowleft$ $\times k$ merges};
  \end{scope}
  \draw[line, line width=.6pt] (.5,1.5) -- (6.1,1.5);
  \node[font=\sffamily\scriptsize\bfseries, text=ftc, anchor=west] at (.5,1.3) {OBJECTIVE};
  \node[base] at (3.3,.8) {$\displaystyle\min_{\tilde\theta}\ \mathbb{E}\,\big\|\tilde f_\ell(h_\ell)-h_{\ell+2}\big\|_2^2$};
  \node[note] at (3.3,.26) {no labels \ $\cdot$ \ no new modules \ $\cdot$ \ 300 local steps};
  \end{scope}

  \begin{scope}[xshift=14.6cm]
  \node[card, anchor=south west, minimum width=4.6cm, minimum height=\H cm] at (0,0) {};
  \begin{scope}[yshift=.6cm]
  \node[badge] at (.55,7.25) {3};
  \node[ptitle] at (.95,7.25) {Deploy $L{-}k$ Model};
  \node[note, anchor=west] at (.95,6.8) {standard transformer, $k{=}1$ shown};

  \foreach \i/\nm in {0/{f_{\ell+3}}, 1/{f_{\ell+2}}, 3/{f_{\ell-1}}, 4/{f_{\ell-2}}} {
    \pgfmathsetmacro{\y}{5.55-0.78*\i}
    \node[gblk] at (2.0,\y) {$\nm$};
  }
  \node[blk, fill=ftc!20, draw=ftc, text=ftc!55!black, line width=1.3pt] at (2.0,3.99) {$\tilde f_{\ell}$};
  \draw[flow] (2.0,1.55) -- (2.0,2.03);
  \draw[flow] (2.0,5.9) -- (2.0,6.27);
  \draw[decorate, decoration={brace, amplitude=4pt, mirror}, draw=slate, line width=.7pt]
    (3.3,2.1) -- (3.3,5.88);
  \node[note, rotate=90, anchor=north, text=ink] at (3.62,3.99) {$L{-}k$ layers};

  \end{scope}
  \draw[line, line width=.6pt] (.5,1.5) -- (4.1,1.5);
  \node[font=\sffamily\scriptsize\bfseries, text=ink, anchor=west] at (.5,1.3) {RESULT};
  \node[base] at (2.3,.8) {$\approx L/(L{-}k)\times$ faster decode};
  \node[note] at (2.3,.26) {\textbf{0} new inference parameters};
  \end{scope}

  \draw[big] (6.5,4.35) -- (7.1,4.35);
  \draw[big] (13.9,4.35) -- (14.5,4.35);

  \end{tikzpicture}%
  }
  \caption{\XMerge in three stages. \textbf{(1)~Select.} Each block is scored on its own hidden-state
  transition $h_\ell\!\to\!h_{\ell+1}$ by relative-magnitude ($\mathrm{RM}$, magnitude change) and
  Block-Influence ($\mathrm{BI}$, angular change) from WikiText-2 activations; the two axes are $z$-scored
  and fused by a maximum, $s_\ell=\max(z^{\mathrm{RM}}_\ell,z^{\mathrm{BI}}_\ell)$, and the block with the
  smallest fused score is removed, so a block is chosen only when it is quiet on \emph{both} axes; one
  with low magnitude but high angular change is rejected. \textbf{(2)~Merge.} The selected adjacent pair
  $(f_\ell,f_{\ell+1})$ is collapsed into a single surviving standard block $\tilde f_\ell$ by a label-free,
  300-step local fit that minimizes the mean-squared error between the pair's frozen output $h_{\ell+2}$ and
  the merged block's output $\tilde h_{\ell+2}$ (no labels, no new modules); the process repeats for each of
  the $k$ merges. \textbf{(3)~Deploy.} The result is a standard $L{-}k$ transformer ($k{=}1$ shown) with zero
  added inference-time parameters and $\approx L/(L{-}k)\times$ faster decode. Definitions in
  \S\ref{sec:selector-def}--\S\ref{sec:operator}.}
  \label{fig:overview}
\end{figure*}

\subsection{Depth Compression as Select-then-Merge}\label{sec:formulation}
Figure~\ref{fig:overview} provides an overview of \XMerge. A transformer applies blocks $f_1,\dots,f_L$, with block $f_l$
mapping $h_l\!\to\!h_{l+1}$. The selector $\mathcal{S}$ scores each single-block transition and marks a block
$f_\ell$ to remove; the operator $\mathcal{M}$ then replaces the adjacent pair $(f_\ell,f_{\ell+1})$ (with
original mapping $h_{\ell+2}=f_{\ell+1}(f_\ell(h_\ell))$) by one standard block $\tilde f_\ell$ giving
$\tilde h_{\ell+2}=\tilde f_\ell(h_\ell)$, so $\mathcal{C}=\mathcal{M}\circ\mathcal{S}$. We study the two
failure modes (\S\ref{sec:intro}) separately: the selector in \S\ref{sec:fusion}, the operator in
\S\ref{sec:loadbearing}. After $k$ merges the model has $L-k$ blocks, a shape shared by all methods at the
same $k$.

\subsection{Selector: Cross-Axis Fusion (Parameter-Free)}\label{sec:selector-def}
\paragraph{Relative-magnitude axis.}
Low $\RM$ = small residual displacement; high $\RM$ = the layer strongly changes activation magnitude.
\[
  \RM_l = \mathbb{E}_{x,t}\!\left[\;\dfrac{\lVert h^{(x,t)}_{l+1}-h^{(x,t)}_{l}\rVert_2}
  {\lVert h^{(x,t)}_{l}\rVert_2+\epsilon}\;\right].
\]

\paragraph{Angular axis.}
Low $\BI$ = little directional change; high $\BI$ = the layer rotates the representation substantially.
\[
  \BI_l = \mathbb{E}_{x,t}\!\left[\;1-\dfrac{\langle h^{(x,t)}_{l},h^{(x,t)}_{l+1}\rangle}
  {\lVert h^{(x,t)}_{l}\rVert_2\,\lVert h^{(x,t)}_{l+1}\rVert_2+\epsilon}\;\right].
\]
This is the Block-Influence criterion of ShortGPT~\cite{men2024shortgpt}. We reuse it as the angular axis and
combine it with relative magnitude. The reconstruction operator is the main methodological contribution; the
cross-axis combination is used as a safeguard when the two axes disagree.

\paragraph{Cross-axis score.} Both axes score a single block $f_l$ on its input$\to$output transition.
Standardize each across eligible layers,
$\zRM_l=(\RM_l-\mu_{\RM})/(\sigma_{\RM}+\epsilon)$ and $\zBI_l=(\BI_l-\mu_{\BI})/(\sigma_{\BI}+\epsilon)$,
and mark for removal the eligible block with the smallest
\[
  s_l=\max\!\left(\zRM_l,\;\zBI_l\right).
\]
A block is quiet only when \emph{neither} magnitude nor angular change is large. We use $\max$ rather than an average
because an average can hide a large score on one axis with a small score on the other. This choice has no
tunable weighting coefficient. No boundary mask is used in the reported runs: an optional
safeguard forbidding the first/last blocks as merge \emph{targets} (motivated by the boundary sensitivity of
layer surgery~\cite{quantdamage2026}) is \emph{off} for every reported cell, since cross-axis fusion already
avoids the boundary selections that would trigger it (\S\ref{sec:fusion}); we disclose it only to note it was
not needed.

\subsection{Operator: Reconstructive Depth Collapse}\label{sec:operator}
Local activation-matching reconstruction after pruning is itself an established paradigm
(\S\ref{sec:related}); our novelty is \emph{not} local reconstruction \emph{per se} but its serving-preserving
instantiation. We distill the \emph{adjacent two-block mapping into one of the existing surviving standard
blocks} rather than a learned replacement module, so the compressed model remains an ordinary shallower
transformer with no inserted module.
For calibration inputs $x$, we cache the pair's input $h_l(x)$ and original output $h_{l+2}(x)$, and retain one
member of the pair as the surviving block $\tilde f_l$: the neighbor on the side with the higher adjacent
activation-patching redundancy score $S_{\mathrm{patch}}$ when both sides are available, initialized from its
own pretrained weights. This fixed rule sets only the merge direction; we do not treat it as a contribution or
ablate it (\S\ref{app:selector}, \S\ref{sec:limits}). We then optimize \emph{all} of its parameters (attention, MLP, and norms)
in place:
\[
  \min_{\tilde\theta_l}\;\mathbb{E}_x\!\left[\;
  \bigl\lVert \tilde f_l\!\left(h_l(x);\tilde\theta_l\right)-f_{l+1}\!\left(f_l(h_l(x))\right)\bigr\rVert_2^2
  \;\right].
\]
Optimization is Adam for 300 steps, learning rate $10^{-5}$, batch size 16, over 128 WikiText-2 \emph{train}
sequences of 512 tokens, the \emph{same} configuration for every model (not per-model tuned), \textbf{without
labels, downstream supervision, end-to-end optimization, or new inference parameters}. We call
this \emph{layer-local post-training reconstruction} (not ``fine-tuning-free''). Turning reconstruction off
($\text{steps}{=}0$) collapses \XMerge to a selected drop that falls \emph{below} baselines
(\S\ref{sec:loadbearing}): the gain comes from reconstruction, not selection.

\subsection{Iterative Compression}\label{sec:iterative}
For $k>1$, selection order is computed \emph{once} from the original model, while reconstruction targets are
refreshed after each merge from the current partially compressed model. Candidates are processed in
ascending fused-score order; a candidate that has already been removed, or whose neighbors have both already
been removed, is skipped in favor of the next eligible candidate.

\section{Experimental Setup}\label{sec:setup}

\begin{table}[htbp]
  \centering\small
  \caption{Backbones (two families, 0.5B--8B): Llama-3~\cite{dubey2024llama3},
  Qwen2.5~\cite{yang2024qwen25}, and Qwen3~\cite{qwen2025qwen3}. Depth-reduction \% shown for $k{=}1/2/4$.}
  \label{tab:models}
  \begin{tabular}{lccccc}
    \toprule
    Model & Layers & Dense CORE & Dense MMLU & Dense PPL & \%\,removed ($k{=}1/2/4$) \\
    \midrule
    Llama-3.2-1B & 16 & 0.366 & 0.352 & 8.59  & 6.3 / 12.5 / 25.0 \\
    Llama-3.2-3B & 28 & 0.496 & 0.405 & 6.91  & 3.6 / 7.1 / 14.3 \\
    Llama-3-8B   & 32 & 0.561 & 0.467 & 7.33  & 3.1 / 6.3 / 12.5 \\
    Qwen2.5-0.5B & 24 & 0.322 & 0.336 & 11.50 & 4.2 / 8.3 / 16.7 \\
    Qwen3-0.6B   & 28 & 0.304 & 0.320 & 18.29 & 3.6 / 7.1 / 14.3 \\
    Qwen3-1.7B   & 28 & 0.433 & 0.379 & 15.05 & 3.6 / 7.1 / 14.3 \\
    Qwen3-8B     & 36 & 0.417 & 0.336 & 8.58  & 2.8 / 5.6 / 11.1 \\
    \bottomrule
  \end{tabular}
\end{table}

\paragraph{Baselines and comparison protocol.} We compare against five published methods (\emph{dropping:}
ShortGPT~\cite{men2024shortgpt}; \emph{merging/collapsing:} LaCo~\cite{yang2024laco}, MKA~\cite{mka2024},
SWM~\cite{swm2024}, CoMe~\cite{come2024}) under two settings. \emph{Setting~A}
(\S\ref{sec:results}), the \emph{constructed-checkpoint operator comparison}, has every method produce the same
$L{-}k$ architecture using its authors' native selection and operator at matched depth, with \emph{no}
follow-on recovery. It fixes the \emph{deployed artifact} (identical architecture, parameter count,
latency), not construction compute: \XMerge's operator does gradient work the training-free baselines skip, so
its checkpoint costs minutes--hours rather than seconds--minutes (break-even in \S\ref{app:cost};
\S\ref{sec:limits}). We assign \XMerge's fast, layer-local,
label-free reconstruction to the operator (like the analytic fold in LaCo/SWM) and include it here,
whereas CoMe's hierarchical distillation and SWM's LoRA are global post-hoc stages deferred to \emph{Setting~B}
(\S\ref{sec:recovery}), where every \emph{healed} method receives the \emph{same} budget (four methods have
recovery checkpoints; LaCo/MKA are analytic folds we did not heal, Setting~A only). All six otherwise share the
same $L{-}k$ architecture, zero inference-time parameters, and label-free inputs; baseline deviations are
audited in \S\ref{app:audit}. We do \emph{not} run SLEB~\cite{song2024sleb} (whole-block drop, ShortGPT's
family) or BlockPruner~\cite{blockpruner2024} (finer sub-block granularity, a different axis); both are cited
as related work.

\paragraph{Compression levels.} $k\in\{1,2,4\}$ absolute removed-layer counts (mild$\to$aggressive);
percentage reductions in Table~\ref{tab:models}.
\paragraph{Metrics.} Our primary metric is CORE~\cite{li2024datacomplm}, a 22-task centered aggregate (scored
with the nanochat bundle~\cite{nanochat}) covering zero-shot and in-context-learning (ICL) tasks. We use it
because it is more discriminative in this study than MMLU (Massive Multitask Language Understanding) and,
unlike perplexity, its task scores are less directly tied to the reconstruction corpus. We additionally report zero-shot MMLU~\cite{hendrycks2021mmlu} and
WikiText-2~\cite{merity2017wikitext} test perplexity (PPL); reconstruction calibrates on WikiText-2 train, so
perplexity is partially in-domain despite disjoint train/test splits. We prespecify a \emph{collapse}
threshold (a model/regime centered CORE below $0.10$, near-random) for the regime analysis
(\S\ref{sec:regime}). Full task composition, prompting, centering, and evaluation details are in
\S\ref{app:regime} and \S\ref{app:eval}.
\paragraph{Protocol.} Single compression seed 42; reconstruction hyperparameters (HPs) are fixed across all models
(\S\ref{sec:operator}) and evaluation runs in \texttt{float16}. All $7\times6\times3=126$ backbone cells are
complete; hardware and evaluation-sharding details are in \S\ref{app:eval}.

\section{Main Results}\label{sec:results}

We report the most aggressive setting ($k{=}4$) in the main text and defer the full
$k{\in}\{1,2,4\}$ six-method grids for every metric to \S\ref{app:core}--\S\ref{app:ppl}.

\begin{figure}[htbp]
  \centering
  \includegraphics[width=0.72\linewidth]{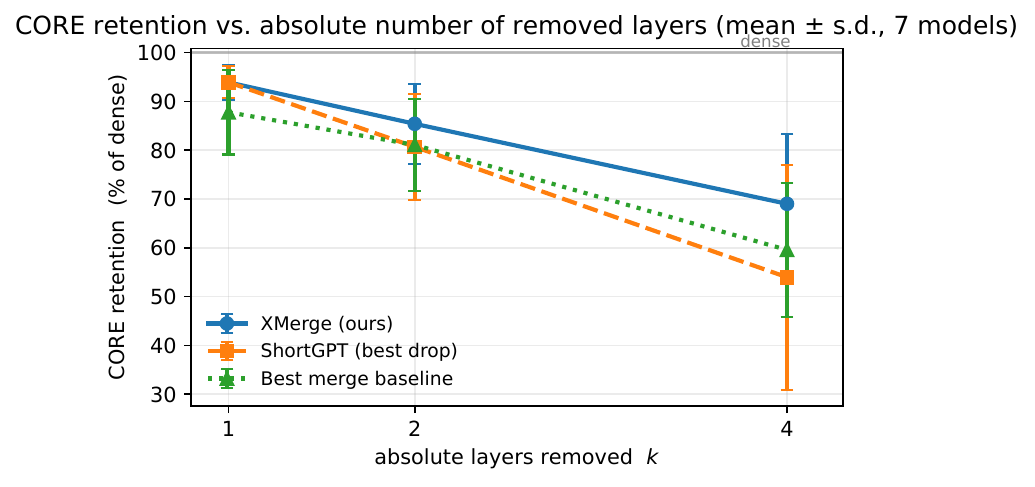}
  \caption{Mean CORE retention (compressed/dense) as the \emph{absolute} number of removed layers grows from
  $k{=}1$ to $4$ (mean$\pm$s.d.\ over 7 backbones). A fixed $k$ is a different relative reduction per backbone,
  so this shows robustness to larger absolute removal, not a matched-ratio scaling law. \XMerge vs.\ best drop
  (ShortGPT) and best competing merge baseline (per-$(\text{model},k)$ max over LaCo/MKA/SWM/CoMe).}
  \label{fig:retention}
\end{figure}

\begin{table}[htbp]
  \centering\small
  \caption{Main results at $k{=}4$: \XMerge (XM) vs.\ the strongest per-model baseline (``bb'') on each metric,
  with the winning baseline named for CORE and MMLU (full grids \S\ref{app:core}--\S\ref{app:ppl}). Among
  baselines LaCo has the lowest PPL in every row, so each wiki-PPL bb is LaCo (not the overall best; \XMerge
  beats LaCo's PPL on six of seven). CORE$\uparrow$, MMLU$\uparrow$, wiki-PPL$\downarrow$; bold = winner.}
  \label{tab:main}
  \begin{tabular}{lccc}
    \toprule
    Model & CORE (XM / bb) & MMLU (XM / bb) & wiki-PPL (XM / bb; bb = LaCo) \\
    \midrule
    Llama-3.2-1B & \textbf{.201} / .130 (MKA)      & \textbf{.297} / .289 (ShortGPT) & \textbf{27.16} / 48.18 \\
    Llama-3.2-3B & \textbf{.406} / .373 (ShortGPT) & \textbf{.374} / .359 (SWM)      & \textbf{10.49} / 14.53 \\
    Llama-3-8B   & \textbf{.534} / .528 (ShortGPT) & \textbf{.434} / .427 (ShortGPT) & \textbf{9.55} / 10.60 \\
    Qwen2.5-0.5B & \textbf{.178} / .168 (ShortGPT) & \textbf{.297} / .296 (SWM)      & \textbf{13.94} / 16.79 \\
    Qwen3-0.6B   & \textbf{.179} / .165 (MKA)      & .295 / \textbf{.296} (LaCo)     & \textbf{20.44} / 25.88 \\
    Qwen3-1.7B   & .271 / \textbf{.289} (MKA)      & \textbf{.338} / .322 (MKA)      & \textbf{16.33} / 20.19 \\
    Qwen3-8B     & \textbf{.310} / .262 (SWM)      & \textbf{.314} / .308 (LaCo)     & 12.89 / \textbf{11.17} \\
    \bottomrule
  \end{tabular}
\end{table}
At $k{=}4$, \XMerge is strongest on six of seven backbones on CORE and, separately, on six of seven on MMLU.
The exception differs by metric, so \XMerge is best on both CORE and MMLU for five of seven backbones, and no
backbone is an exception on more than one metric (Table~\ref{tab:main}). The CORE win count
rises with the absolute number of removed layers (four, five, six of seven at $k{=}1,2,4$): at $k{=}1$ the
picture is mixed, with \XMerge sole-best on only three backbones and tying or trailing ShortGPT elsewhere, so
simpler methods are competitive at mild compression and the advantage becomes decisive only under aggressive
removal (full grids in \S\ref{app:core}--\S\ref{app:ppl}; retention trend in Fig.~\ref{fig:retention}). The
three $k{=}4$ exceptions are narrow (Qwen3-1.7B CORE, \S\ref{sec:limits}; Qwen3-0.6B MMLU by $0.001$;
Qwen3-8B PPL). We report all perplexity values without truncation and impose no numerical collapse threshold. Only \XMerge and LaCo
avoid the extreme $k{=}4$ blow-ups of the other four ($10^2$--$10^4$; e.g.\ MKA $8343$ on Llama-3.2-1B, CoMe
$5.3{\times}10^4$ on Qwen3-1.7B; \S\ref{app:ppl}), and \XMerge is lower than LaCo on six of seven. MKA shows
the risk directly: competitive CORE on Qwen3-1.7B but perplexity $440$, so reporting accuracy and perplexity
together reveals the degradation.

\paragraph{Signal vs.\ task noise.} CORE is a 22-task macro-average, so we can bound how much of each $k{=}4$
margin is separable from task-level sampling noise by resampling the 22 per-task scores with replacement
(paired $20{,}000$-sample bootstrap over XMerge minus the strongest per-model baseline;
Table~\ref{tab:bootstrap}, \S\ref{app:bootstrap}).
The three largest margins have 95\% CIs excluding zero (Llama-3.2-1B $+.070$, Llama-3.2-3B $+.033$,
Qwen3-8B $+.048$), while the three narrow wins (Llama-3-8B $+.006$, Qwen2.5-0.5B $+.010$, Qwen3-0.6B $+.013$)
and the single Qwen3-1.7B loss ($-.018$) have CIs spanning zero and should be read as ties. The one-sided
sign test for the six-of-seven pattern gives $p{=}0.06$, so the cross-backbone count is not significant at
conventional levels. This bootstrap measures task-sampling uncertainty only; it does not estimate
reconstruction-seed variance, which we do not measure (\S\ref{sec:limits}).

At $k{=}1$ \XMerge and ShortGPT each retain $\approx$94\% of dense CORE, while the best competing merge
baseline retains only $\approx$88\%; the gap is largest under aggressive removal: at $k{=}4$ \XMerge holds
\textbf{69\%} of dense CORE, versus \textbf{60\%} for the best competing merge baseline and \textbf{54\%} for
ShortGPT (Fig.~\ref{fig:retention}). The margin over the best merge baseline does not grow monotonically
($+6.1$, $+4.2$, $+9.5$ points at $k{=}1,2,4$).

\begin{figure*}[t]
  \centering
  \includegraphics[width=\linewidth]{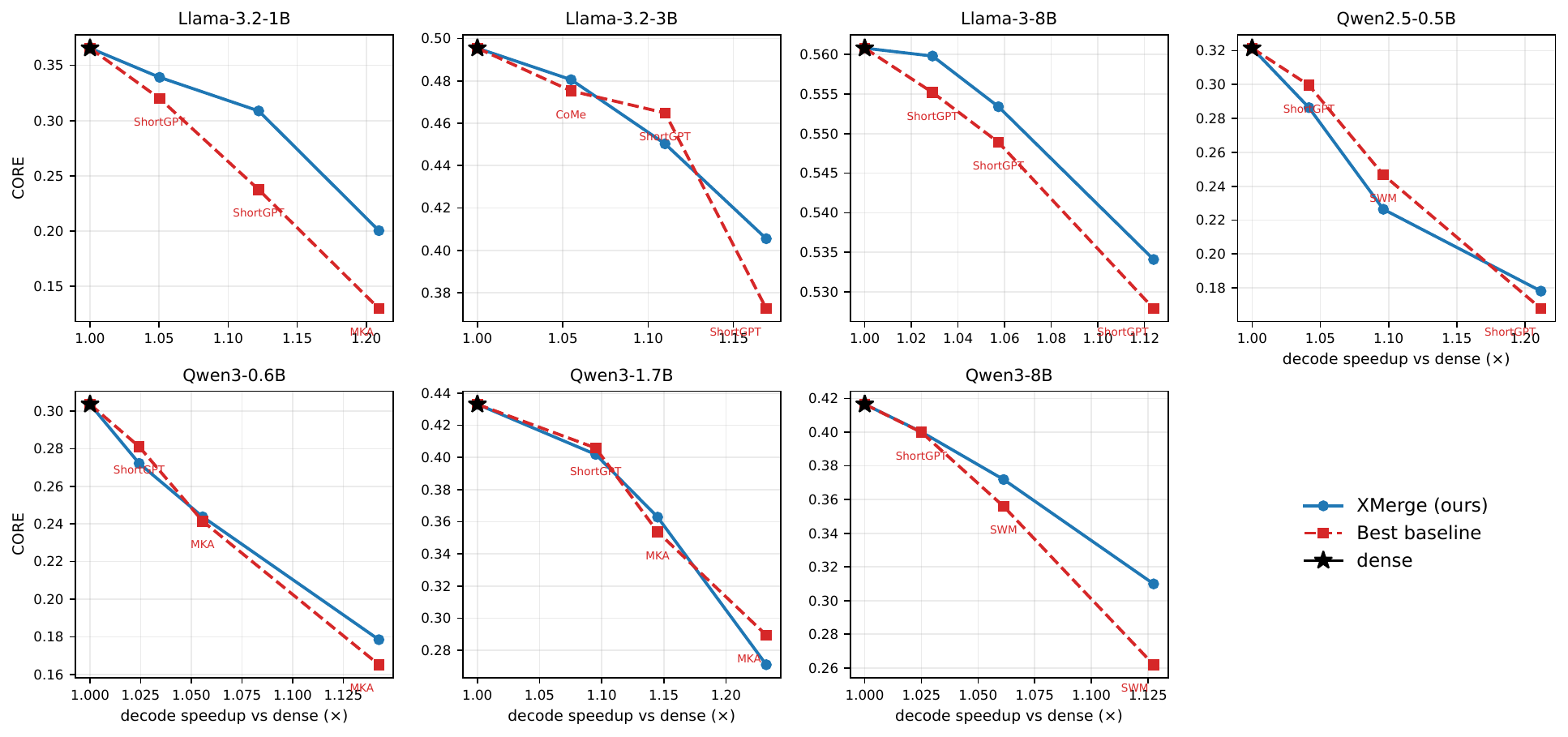}
  \caption{Quality--latency Pareto per backbone (Setting~A, operators only, no recovery): CORE vs.\ \emph{measured}
  batch-1 decode speedup (\(\star\)=dense; points $k{=}1,2,4$). All operators share the same $L{-}k$
  architecture at matched speedup, so higher-and-to-the-right is better; \XMerge is best at the $k{=}4$ endpoint on
  six of seven backbones (best-or-tied 15/21; trails at $k{=}1{,}2$ on Qwen2.5-0.5B). Protocol + speedup table
  \S\ref{app:latency}.}
  \label{fig:pareto}
\end{figure*}

\paragraph{Quality at matched inference cost.} Because a fixed $k$ yields the identical $L{-}k$ decoder for
every operator, the accuracy differences above are obtained at \emph{identical} inference cost. We verify this
directly by benchmarking decode throughput on the real checkpoints: measured batch-1 speedup is
depth-proportional and method-independent ($\approx L/(L{-}k)$; mean $1.04/1.09/1.17\times$ at $k{=}1/2/4$,
full protocol and per-backbone table in \S\ref{app:latency}). The quality--latency Pareto frontier is thus
set entirely by which operator retains the most quality per removed layer: \XMerge is best at the $k{=}4$ endpoint
on six of seven backbones and is best-or-tied in 15/21 matched-depth CORE cells (Fig.~\ref{fig:pareto}).

\section{Component Analysis}\label{sec:analysis}
We next separate the contributions of the operator (\S\ref{sec:loadbearing}) and the selector
(\S\ref{sec:fusion}) and evaluate robustness across prompting regimes (\S\ref{sec:regime}).

\subsection{Reconstruction Provides Most of the Gain}\label{sec:loadbearing}
With selection fixed, reconstruction raises the score above the published baselines, whereas the corresponding
selected drop is below them. In our two-backbone on/off ablation at $k{=}4$
under RM selection, Llama-3.2-3B improves from .344 (drop) to .406 (reconstruct) and Qwen3-0.6B from .136 to
.184; in both cases $\text{selection+drop} < \text{published baselines} < \text{selection+reconstruction}$
(we did not run the full seven-model grid).
On Llama-3.2-3B, RM and max-fusion select the same layers, so .406 is also the deployed \XMerge number; on
Qwen3-0.6B, deployed max-fusion is .179, trading a small CORE decrease for better perplexity and MMLU
(\S\ref{sec:fusion}). These results indicate that the gain does not come from selection alone.

\subsection{Cross-Axis Fusion Reduces Disagreement Risk}\label{sec:fusion}
The selector combines RM and BI with $\max$ (\S\ref{sec:selector-def}), fixed a priori and applied identically
to all 7 models. An oracle backtest confirms RM and BI are the two informative axes and are near-tied as
individual CORE predictors (\S\ref{app:selector}), so fusion is adopted for robustness, not a claim that either
axis is superior. Because $\max$ preserves the single-axis choice when
the two axes agree, fusion is byte-identical to RM on 5/7 backbones and alters the
decision only on Qwen3-0.6B and Qwen3-1.7B, where the two axes are less correlated
($\mathrm{corr}(\RM,\BI)\!\approx\!0.01$ where it acts vs.\ $0.74$ where it is a no-op; Table~\ref{tab:selfimit},
\S\ref{app:selector}). In these two cases, each single axis chooses a layer set that leads to substantially worse results, which
fusion avoids: RM alone drops \emph{early} layers on Qwen3-1.7B (\{3,4,5,7\}), while BI alone selects the
\emph{final} block and collapses the model (Qwen3-1.7B PPL $5.1{\times}10^4$ under reconstruction; Qwen3-0.6B
PPL $226$ under a plain drop), as the head-to-head at $k{=}4$ shows (Table~\ref{tab:selector-h2h}).

\begin{table}[htbp]
  \centering\small
  \caption{Selector head-to-head at $k{=}4$ (CORE / wiki-PPL / MMLU), on the two backbones where cross-axis
  fusion overrides RM; identical reconstruction HPs, so only selection differs (max-fusion adopted). BI-only
  collapses (Qwen3-1.7B PPL $5.1{\times}10^4$, selecting the final block; Qwen3-0.6B under a plain drop, CORE
  $.079$ / PPL $226$); MKA shown for reference (high PPL despite competitive CORE). Narration in
  \S\ref{sec:fusion}.}
  \label{tab:selector-h2h}
  \begin{tabular}{lcc}
    \toprule
    selector & Qwen3-1.7B (dense .433) & Qwen3-0.6B (dense .304) \\
    \midrule
    RM-only & .205 / 19.37 / .303 & \textbf{.184} / 21.13 / .294 \\
    BI-only & .063 / 50606 / --- & --- (drop: .079 / 226) \\
    \textbf{max-fusion (RM$\oplus$BI)} & \textbf{.271} / \textbf{16.33} / \textbf{.338} & .179 / \textbf{20.44} / \textbf{.295} \\
    MKA & .289 / 440 / .322 & .165 / 276 / .294 \\
    \bottomrule
  \end{tabular}
\end{table}

On Qwen3-0.6B the fused choice trades a small CORE decrease for better perplexity and MMLU, improving
worst-case behavior rather than winning every metric. We therefore treat reconstruction as the primary operator
contribution and cross-axis fusion as the robustness mechanism, since a strong operator alone
cannot compensate for an unsafe structural choice.

\subsection{Zero-Shot vs.\ In-Context Regimes}\label{sec:regime}
Because CORE aggregates zero-shot and in-context-learning tasks, we split it into the two regimes and ask a
robustness question: does each operator preserve \emph{both}? Using the prespecified collapse threshold
(\S\ref{sec:setup}; centered CORE below $0.10$), we count collapses across all 14 cells (7 models $\times$ 2
regimes). \XMerge is the only \emph{evaluated} operator that never collapses: zero below-threshold cells and
top-2 in all 7 models in \emph{both} regimes (worst cell $+0.172$; Table~\ref{tab:breaks}). Every baseline
collapses on at least one cell (CoMe on 7 of 14, with MKA and CoMe going \emph{negative} at ICL on
Llama-3.2-1B), so \XMerge's clearest distinction from the baselines is this zero-collapse robustness across
regimes, not any single win.

\begin{table}[htbp]
  \centering\small
  \caption{Robustness across all 14 (model $\times$ regime) cells at $k{=}4$. A break is centered CORE
  $<0.10$; ``floor'' is each operator's worst cell. Per-model 0-shot/ICL tables are in \S\ref{app:regime}.}
  \label{tab:breaks}
  \begin{tabular}{lcccc}
    \toprule
    Method & breaks (of 14) & floor (min) & top-2 @ 0-shot & top-2 @ ICL \\
    \midrule
    \textbf{XM (\XMerge)} & \textbf{0} & \textbf{+.172} & \textbf{7/7} & \textbf{7/7} \\
    SWM      & 1 & +.093 & 0/7 & 2/7 \\
    LaCo     & 2 & +.064 & 2/7 & 0/7 \\
    MKA      & 2 (incl.\ neg.) & $-$.022 & 1/7 & 2/7 \\
    ShortGPT & 3 & +.071 & 4/7 & 3/7 \\
    CoMe     & 7 (incl.\ neg.) & $-$.025 & 0/7 & 0/7 \\
    \bottomrule
  \end{tabular}
\end{table}

Within each regime, \XMerge has the highest mean score (0-shot mean $.291$ vs.\ next-best ShortGPT $.236$; ICL $.300$ vs.\
$.242$; per-model means and the full per-regime tables in \S\ref{app:regime}). The Qwen3-1.7B/0.6B exception
(\S\ref{sec:limits}) is \emph{purely} an ICL effect: \XMerge wins the 0-shot slice on both.

\section{Recoverability Under a Shared Post-Compression Budget}\label{sec:recovery}

We next test whether \XMerge also provides a strong starting point for post-compression recovery.
\emph{Setting~B} grants every included method the same additional healing budget: one uniform
WikiText-2 knowledge-distillation (KD) stage with identical data, LoRA rank, learning-rate (LR) policy, teacher, and wall-clock budget, applied to each compressed model, isolating which operator gives the best starting point. This
controls the recovery stage, not total compression-plus-recovery compute (native construction costs differ,
\S\ref{sec:limits}). We heal the four methods with recovery checkpoints (\XMerge, ShortGPT, SWM, CoMe); the
analytic folds LaCo/MKA appear in Setting~A only.

\begin{table}[htbp]
  \centering\small
  \caption{Recovery under a shared \emph{additional} post-compression budget ($k{=}4$), mean over the LR grid
  $\{1,2,5\}\times10^{-5}$ (no LR tuned on the eval metric; per-LR + full protocol \S\ref{app:recovery}). The
  four healed methods each get the same follow-on WikiText-2 KD stage on their own compressed model (LaCo/MKA
  are Setting-A only). Bold = best per metric (CORE, PPL scored separately) within a model; a merge baseline has the lowest
  raw PPL at each scale (SWM@1B, CoMe@8B).}
  \label{tab:recovery}
  \begin{tabular}{lcccc}
    \toprule
    Model (dense CORE / PPL) & \XMerge & ShortGPT & CoMe & SWM \\
    \midrule
    Llama-3.2-1B (.366 / 8.59) & \textbf{.252} / 15.43 & .245 / 16.56 & .170 / 17.11 & .223 / \textbf{15.29} \\
    Qwen3-8B (.417 / 8.58)     & \textbf{.358} / 10.66 & .351 / 11.57 & .343 / \textbf{9.23} & \textbf{.358} / 11.89 \\
    \bottomrule
  \end{tabular}
\end{table}
Under the shared follow-on budget, \XMerge starts from the strongest unhealed model at both scales. It attains
the best recovered CORE on Llama-3.2-1B ($+0.007$ over the next method) and ties SWM on Qwen3-8B ($.358$ vs.\
$.358$; unrounded gap $\approx0.0002$, within single-seed noise), while keeping the best perplexity among the
CORE-competitive methods; its advantage is thus not limited to the unhealed checkpoint.

\section{Discussion and Limitations}\label{sec:limits}
\paragraph{Where \XMerge wins.} The Setting~A per-cell winners show which baselines it competes with: the strongest single
baseline is most often ShortGPT (12/21 (model,$k$) cells; MKA 5, SWM 3, CoMe 1, LaCo 0), expected since
Setting~A withholds the mergers' native LoRA/distillation (deferred to Setting~B, \S\ref{sec:recovery}) and favors
pure droppers. \XMerge exceeds the strongest baseline in 15/21 cells using only label-free
reconstruction, and under Setting~B's shared budget stays best-or-tied (a win at 1B, a tie with SWM at 8B).

\paragraph{Reliability checks.} A compressed model is only useful if it stays reliable, not just fast, so we
evaluate two aspects explicitly. \emph{Task-ability robustness:} \XMerge is the only \emph{evaluated} operator
with no CORE collapse across all 14 (model, regime) cells (\S\ref{sec:regime}) and no perplexity blow-up at
$k{=}4$ (\S\ref{sec:results}). \emph{Calibration:} because
the operator matches dense hidden states, it should distort confidence less than analytic merges; we test this
with expected calibration error (ECE) on MMLU. On Llama-3-8B at $k{=}4$ (Setting~A), \XMerge has the
\emph{lowest} ECE degradation of all six operators ($\Delta\mathrm{ECE}{=}{+}0.010$ over the dense model's
$0.125$), better than the analytic weight-fold/merge baselines (LaCo ${+}0.021$, MKA ${+}0.110$) and every
other operator (\S\ref{app:ece}, Table~\ref{tab:ece}, Fig.~\ref{fig:ece}); all methods stay somewhat
overconfident, so reconstruction preserves calibration best but does not fully restore it. This result is limited to one backbone
and one metric; it does not establish a general safety guarantee. We do \emph{not} evaluate the remaining
safety-oriented aspects (hallucination, refusal/safety behaviour), which require instruction- or safety-tuned
variants rather than the base models studied here, and we leave them to future work.

\paragraph{Limitations.} The additional cost is incurred during construction; serving uses the same
standard $L{-}k$ decoder as the baselines, whose decode speedup is a single per-$(\text{backbone},k)$ quantity shared by all
operators (\S\ref{app:latency}), so the gradient work \XMerge does at build time buys quality, not speed. In absolute terms, this one-time cost is
minutes to $4.4$ hours ($\approx$linear in $k$; \S\ref{app:cost}) and is recovered by the per-token saving after
$\approx$$1.9$k--$24$k requests (\S\ref{app:cost}); we do \emph{not} claim parity in total
construction-plus-recovery compute with the training-free baselines. The study uses a single compression seed
(42): the task-bootstrap of \S\ref{sec:results} bounds task-sampling noise but not reconstruction-seed variance
across a full re-run, which only a multi-seed study would close. It covers only dense decoder-only Llama/Qwen
models (0.5B--8B; the 8B ceiling reflects our single 40\,GB GPU-partition compute budget, not a limitation of the
method), not mixture-of-experts (MoE), encoder--decoder, multimodal, or state-space. Reconstruction uses WikiText-2, so WikiText-2
test perplexity is partially in-domain (downstream benchmarks are independent). We do not ablate the
merge-direction heuristic ($S_{\mathrm{patch}}$, \S\ref{sec:operator}). \XMerge is also not universally best:
it loses $k{=}4$ CORE to MKA on Qwen3-1.7B, is not lowest-perplexity on Qwen3-8B, and at mild compression
simpler methods can match it; the advantage appears under larger absolute layer removal and in cross-regime
robustness. Because the reconstruction objective minimizes hidden-state MSE to the dense model, CORE, MMLU, and
perplexity all partly measure proximity to that same target, so the wins across three metrics are corroborating
views of one objective rather than fully independent confirmations.

\begin{ack}
\subsection*{Acknowledgments}
We thank Prakhar Mehrotra, Chandramouliswaran V, Avinash Karn, Anindya Moitra, Uma Kona, Angela McAtee, Linsey Pang, and Yun-Shiuan Chuang for their organizational support and coordination throughout this work. Jundong Hu additionally thanks Loga Vinayagam for the opportunity to join the team where this work began.

\subsection*{Author Contributions}
\textbf{Jundong Hu:} Led and carried out the research end to end, including conceptualization, methodology, implementation, experimental design and execution, analysis, and manuscript drafting and revision.

\textbf{Shekar Ramachandran:} Provided supervision, compute resources, and manuscript review.
\end{ack}

\bibliographystyle{plainnat}   
\bibliography{references}

\appendix

\section{Hardware and Evaluation Protocol}\label{app:eval}
All runs use NVIDIA A100-SXM4-80GB GPUs partitioned into Multi-Instance GPU (MIG) slices. Evaluation is sharded across 4 MIG instances via
NCCL/DDP: examples are strided across ranks and all-reduced, numerically identical to single-process
evaluation. Evaluation task subsets use a fixed \texttt{Random(1337)} shuffle. MMLU is zero-shot over all
$14{,}042$ examples across 57 subjects; WikiText-2 perplexity uses a sliding window (max-length 2048, stride
512) on the \emph{test} split, while reconstruction calibrates on the \emph{train} split (disjoint).

\section{Full CORE Tables}\label{app:core}
All six methods $\times$ seven models $\times$ $k\in\{1,2,4\}$; bold = best in row. \XMerge (``XM'') uses
max-fusion selection on Qwen3-0.6B ($k{=}4$) and Qwen3-1.7B (all $k$) and is byte-identical to RM
selection elsewhere; MKA is the fixed top-down port. Dense CORE in Table~\ref{tab:models}.

\begin{table}[htbp]
  \centering\small
  \caption{CORE, $k{=}1$.}
  \begin{tabular}{lcccccc}
    \toprule
    Model & ShortGPT & LaCo & MKA & SWM & CoMe & XM \\
    \midrule
    Llama-3.2-1B & .320 & .254 & .270 & .254 & .266 & \textbf{.339} \\
    Llama-3.2-3B & .472 & .414 & .444 & .401 & .475 & \textbf{.481} \\
    Llama-3-8B   & .555 & .523 & .494 & .523 & .549 & \textbf{.560} \\
    Qwen2.5-0.5B & \textbf{.300} & .240 & .240 & .247 & .232 & .286 \\
    Qwen3-0.6B   & \textbf{.281} & .209 & .260 & .233 & .129 & .272 \\
    Qwen3-1.7B   & \textbf{.406} & .363 & .397 & .359 & .192 & .402 \\
    Qwen3-8B     & .400 & .376 & .380 & .386 & .326 & \textbf{.400} \\
    \bottomrule
  \end{tabular}
\end{table}

\begin{table}[htbp]
  \centering\small
  \caption{CORE, $k{=}2$.}
  \begin{tabular}{lcccccc}
    \toprule
    Model & ShortGPT & LaCo & MKA & SWM & CoMe & XM \\
    \midrule
    Llama-3.2-1B & .238 & .165 & .211 & .227 & .113 & \textbf{.309} \\
    Llama-3.2-3B & \textbf{.465} & .351 & .415 & .391 & .440 & .450 \\
    Llama-3-8B   & .549 & .460 & .469 & .508 & .524 & \textbf{.553} \\
    Qwen2.5-0.5B & .246 & .219 & .184 & \textbf{.247} & .212 & .226 \\
    Qwen3-0.6B   & .239 & .207 & .242 & .198 & .055 & \textbf{.244} \\
    Qwen3-1.7B   & .309 & .293 & .354 & .325 & .070 & \textbf{.363} \\
    Qwen3-8B     & .340 & .343 & .332 & .356 & .213 & \textbf{.372} \\
    \bottomrule
  \end{tabular}
\end{table}

\begin{table}[htbp]
  \centering\small
  \caption{CORE, $k{=}4$.}
  \begin{tabular}{lcccccc}
    \toprule
    Model & ShortGPT & LaCo & MKA & SWM & CoMe & XM \\
    \midrule
    Llama-3.2-1B & .111 & .066 & .130 & .129 & $-$.005 & \textbf{.201} \\
    Llama-3.2-3B & .373 & .212 & .330 & .308 & .279 & \textbf{.406} \\
    Llama-3-8B   & .528 & .358 & .387 & .385 & .458 & \textbf{.534} \\
    Qwen2.5-0.5B & .168 & .158 & .109 & .142 & .095 & \textbf{.178} \\
    Qwen3-0.6B   & .079 & .160 & .165 & .111 & .016 & \textbf{.179} \\
    Qwen3-1.7B   & .167 & .199 & \textbf{.289} & .234 & .010 & .271 \\
    Qwen3-8B     & .257 & .244 & .244 & .262 & .171 & \textbf{.310} \\
    \bottomrule
  \end{tabular}
\end{table}

\section{Task-Bootstrap Confidence Intervals on the CORE Margins}\label{app:bootstrap}
The main sweep uses a single compression seed. To quantify how much of each $k{=}4$ CORE margin is separable
from \emph{task-selection} noise, we treat CORE's 22 constituent per-task scores as the resampling unit: for
each backbone we form the paired per-task differences between \XMerge and the strongest per-model baseline and
draw $20{,}000$ bootstrap resamples of the 22 tasks (with replacement), reporting the mean margin, its 95\%
percentile interval, and $P(\text{margin}>0)$. This is a property of the frozen checkpoints (no new GPU runs)
and is deliberately \emph{not} a seed-variance estimate: it does not perturb the stochastic reconstruction,
so it bounds task-sampling noise only. Reconstruction-seed variance remains unmeasured and is stated as a
limitation (\S\ref{sec:limits}).

\begin{table}[htbp]
  \centering\small
  \caption{Paired task-bootstrap of the $k{=}4$ CORE margin (\XMerge $-$ strongest per-model baseline), $B{=}20{,}000$
  resamples over the 22 CORE tasks. ``(tie)'' marks margins whose 95\% CI includes zero. The three largest
  margins are separable from task noise; the narrow wins and the single Qwen3-1.7B loss are ties. Across
  backbones \XMerge is strongest on six of seven (one-sided sign test $p{=}0.06$). Point columns are rounded
  to match the CORE values in Table~\ref{tab:main} and \S\ref{app:core}; margins and CIs are computed from the
  unrounded per-task scores, so a listed margin can differ by up to $0.001$ from the difference of the
  displayed points.}
  \label{tab:bootstrap}
  \begin{tabular}{lccccc}
    \toprule
    Model & \XMerge & best baseline & margin & 95\% CI & $P(>0)$ \\
    \midrule
    Llama-3.2-1B & .201 & MKA~.130 & $+0.070$ & $[+0.026,\,+0.123]$ & 1.000 \\
    Llama-3.2-3B & .406 & ShortGPT~.373 & $+0.033$ & $[+0.008,\,+0.060]$ & 0.996 \\
    Llama-3-8B   & .534 & ShortGPT~.528 & $+0.006$ & $[-0.004,\,+0.018]$\,(tie) & 0.861 \\
    Qwen2.5-0.5B & .178 & ShortGPT~.168 & $+0.010$ & $[-0.009,\,+0.031]$\,(tie) & 0.842 \\
    Qwen3-0.6B   & .179 & MKA~.165 & $+0.013$ & $[-0.059,\,+0.082]$\,(tie) & 0.654 \\
    Qwen3-1.7B   & .271 & MKA~.289 & $-0.018$ & $[-0.094,\,+0.045]$\,(tie) & 0.326 \\
    Qwen3-8B     & .310 & SWM~.262 & $+0.048$ & $[+0.001,\,+0.110]$ & 0.979 \\
    \bottomrule
  \end{tabular}
\end{table}

\section{Full MMLU Tables}\label{app:mmlu}
Zero-shot MMLU; bold = best in row. MMLU is near the 25\% random floor below 7B (reported for completeness;
discriminative at 8B). Dense MMLU in Table~\ref{tab:models}.

\begin{table}[htbp]
  \centering\small
  \caption{MMLU, $k{=}1$.}
  \begin{tabular}{lcccccc}
    \toprule
    Model & ShortGPT & LaCo & MKA & SWM & CoMe & XM \\
    \midrule
    Llama-3.2-1B & .337 & .322 & .272 & .322 & .325 & \textbf{.340} \\
    Llama-3.2-3B & \textbf{.396} & .380 & .394 & .381 & .390 & .396 \\
    Llama-3-8B   & \textbf{.461} & .448 & .442 & .448 & .453 & .461 \\
    Qwen2.5-0.5B & \textbf{.327} & .310 & .312 & .325 & .307 & .319 \\
    Qwen3-0.6B   & .313 & .311 & .306 & .306 & .290 & \textbf{.319} \\
    Qwen3-1.7B   & .365 & .347 & .361 & .346 & .312 & \textbf{.372} \\
    Qwen3-8B     & .330 & .325 & .327 & .327 & .320 & \textbf{.332} \\
    \bottomrule
  \end{tabular}
\end{table}

\begin{table}[htbp]
  \centering\small
  \caption{MMLU, $k{=}2$.}
  \begin{tabular}{lcccccc}
    \toprule
    Model & ShortGPT & LaCo & MKA & SWM & CoMe & XM \\
    \midrule
    Llama-3.2-1B & .310 & .295 & .272 & .308 & .286 & \textbf{.316} \\
    Llama-3.2-3B & .393 & .348 & .384 & .376 & .372 & \textbf{.395} \\
    Llama-3-8B   & .457 & .412 & .429 & .430 & .438 & \textbf{.460} \\
    Qwen2.5-0.5B & \textbf{.312} & .311 & .301 & .309 & .300 & .308 \\
    Qwen3-0.6B   & .306 & .308 & .304 & .300 & .274 & \textbf{.312} \\
    Qwen3-1.7B   & .341 & .335 & .345 & .334 & .284 & \textbf{.366} \\
    Qwen3-8B     & \textbf{.329} & .321 & .315 & .323 & .294 & .324 \\
    \bottomrule
  \end{tabular}
\end{table}

\begin{table}[htbp]
  \centering\small
  \caption{MMLU, $k{=}4$.}
  \begin{tabular}{lcccccc}
    \toprule
    Model & ShortGPT & LaCo & MKA & SWM & CoMe & XM \\
    \midrule
    Llama-3.2-1B & .289 & .280 & .285 & .280 & .263 & \textbf{.297} \\
    Llama-3.2-3B & .358 & .323 & .352 & .359 & .331 & \textbf{.374} \\
    Llama-3-8B   & .427 & .376 & .411 & .409 & .391 & \textbf{.434} \\
    Qwen2.5-0.5B & .294 & .291 & .281 & .296 & .283 & \textbf{.297} \\
    Qwen3-0.6B   & .271 & \textbf{.296} & .294 & .291 & .261 & .295 \\
    Qwen3-1.7B   & .295 & .310 & .322 & .306 & .269 & \textbf{.338} \\
    Qwen3-8B     & .305 & .308 & .295 & .304 & .306 & \textbf{.314} \\
    \bottomrule
  \end{tabular}
\end{table}

\section{Full Perplexity Tables}\label{app:ppl}
\begin{figure*}[t]
  \centering
  \includegraphics[width=\linewidth]{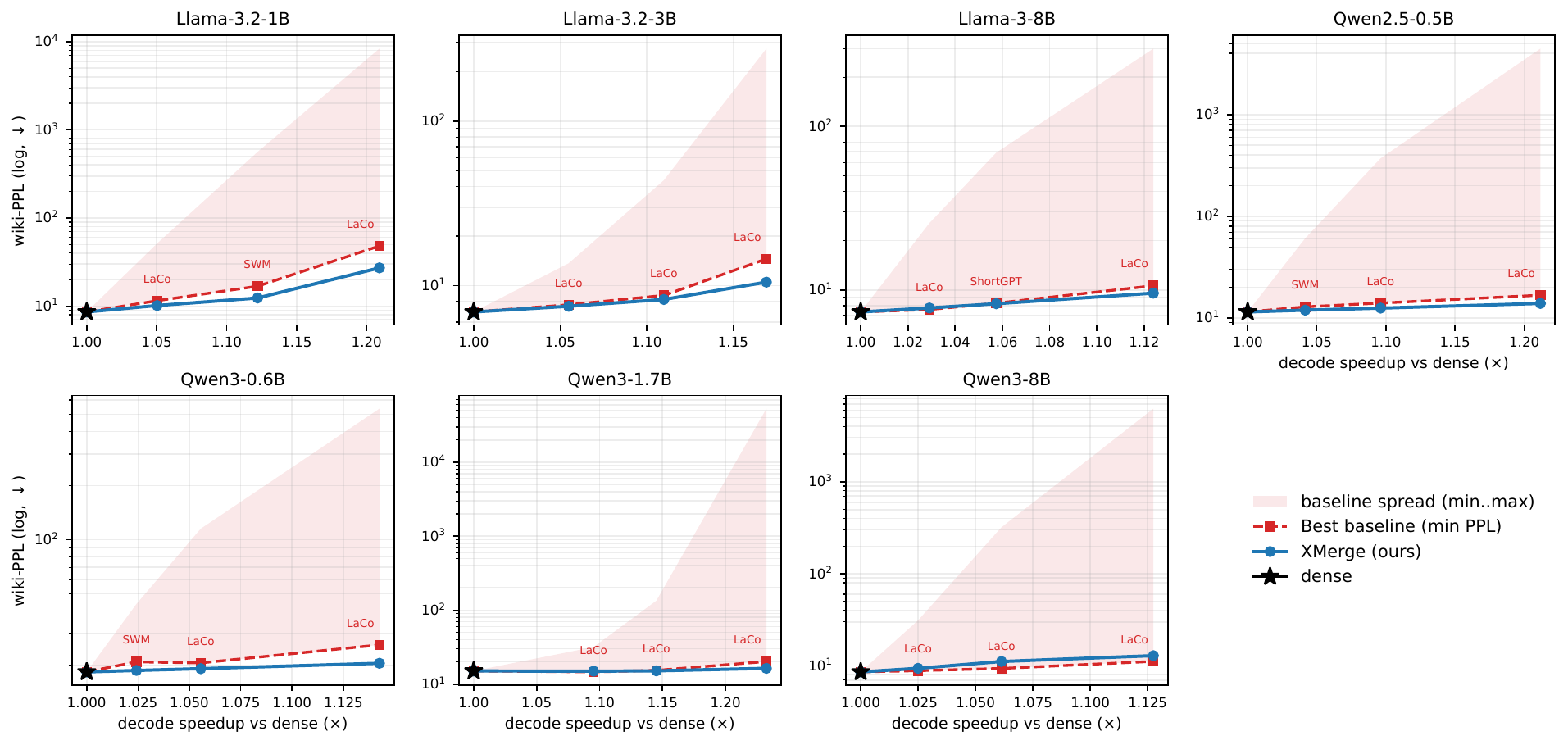}
  \caption{Perplexity vs.\ measured speedup per backbone (Setting~A, recovery-free): WikiText-2 test PPL (log
  scale, lower is better) against \emph{measured} batch-1 decode speedup (\(\star\)=dense at $1\times$; points
  are $k{=}1,2,4$). For the same depth-proportional speedup, \XMerge stays near the dense perplexity while
  several competing operators blow up under aggressive removal (LaCo, like \XMerge, stays bounded; \XMerge is
  lower than LaCo on six of seven backbones at $k{=}4$), the analogous perplexity-versus-speedup plot to the CORE Pareto
  (Fig.~\ref{fig:pareto}). Latency is method-independent at fixed $k$ (\S\ref{app:latency}).}
  \label{fig:ppl_pareto}
\end{figure*}

WikiText-2 test perplexity (lower better); bold = best in row. Extreme values shown untruncated to expose
operator collapses (\XMerge and LaCo are the only operators that avoid the $10^2$--$10^4$ blow-ups at
$k{=}4$; \XMerge is lower than LaCo on six of seven backbones). Dense PPL in Table~\ref{tab:models};
Fig.~\ref{fig:ppl_pareto} plots these values against measured decode speedup.

\begin{table}[htbp]
  \centering\small
  \caption{wiki-PPL, $k{=}1$.}
  \begin{tabular}{lcccccc}
    \toprule
    Model & ShortGPT & LaCo & MKA & SWM & CoMe & XM \\
    \midrule
    Llama-3.2-1B & 12.26 & 11.52 & 51.41 & 11.52 & 21.16 & \textbf{10.16} \\
    Llama-3.2-3B & 7.78 & 7.64 & 13.65 & 8.06 & 8.05 & \textbf{7.48} \\
    Llama-3-8B   & 7.70 & \textbf{7.58} & 25.57 & \textbf{7.58} & 8.14 & 7.75 \\
    Qwen2.5-0.5B & 13.15 & 13.17 & 60.82 & 12.91 & 14.41 & \textbf{11.98} \\
    Qwen3-0.6B   & 22.15 & 21.91 & 43.74 & 20.81 & 26.44 & \textbf{18.64} \\
    Qwen3-1.7B   & 17.93 & \textbf{14.61} & 31.69 & 16.35 & 21.21 & 14.96 \\
    Qwen3-8B     & 9.86 & \textbf{8.85} & 31.14 & 8.90 & 9.14 & 9.39 \\
    \bottomrule
  \end{tabular}
\end{table}

\begin{table}[htbp]
  \centering\small
  \caption{wiki-PPL, $k{=}2$.}
  \begin{tabular}{lcccccc}
    \toprule
    Model & ShortGPT & LaCo & MKA & SWM & CoMe & XM \\
    \midrule
    Llama-3.2-1B & 33.71 & 23.05 & 562 & 16.82 & 198 & \textbf{12.37} \\
    Llama-3.2-3B & 8.87 & 8.71 & 43.68 & 9.02 & 10.47 & \textbf{8.23} \\
    Llama-3-8B   & 8.29 & 8.31 & 69.04 & 8.55 & 9.45 & \textbf{8.23} \\
    Qwen2.5-0.5B & 14.73 & 14.05 & 372 & 15.35 & 16.59 & \textbf{12.54} \\
    Qwen3-0.6B   & 30.94 & 20.53 & 72.22 & 29.86 & 115 & \textbf{19.07} \\
    Qwen3-1.7B   & 31.28 & 15.33 & 82.50 & 18.59 & 134 & \textbf{15.11} \\
    Qwen3-8B     & 11.18 & \textbf{9.36} & 320 & 10.63 & 10.14 & 11.14 \\
    \bottomrule
  \end{tabular}
\end{table}

\begin{table}[htbp]
  \centering\small
  \caption{wiki-PPL, $k{=}4$.}
  \begin{tabular}{lcccccc}
    \toprule
    Model & ShortGPT & LaCo & MKA & SWM & CoMe & XM \\
    \midrule
    Llama-3.2-1B & 403 & 48.18 & 8343 & 1703 & 6726 & \textbf{27.16} \\
    Llama-3.2-3B & 14.91 & 14.53 & 276 & 15.72 & 89.54 & \textbf{10.49} \\
    Llama-3-8B   & 11.92 & 10.60 & 299 & 12.78 & 15.01 & \textbf{9.55} \\
    Qwen2.5-0.5B & 18.54 & 16.79 & 4445 & 31.34 & 58.82 & \textbf{13.94} \\
    Qwen3-0.6B   & 226 & 25.88 & 276 & 80.45 & 537 & \textbf{20.44} \\
    Qwen3-1.7B   & 108 & 20.19 & 440 & 46.83 & 52987 & \textbf{16.33} \\
    Qwen3-8B     & 28.60 & \textbf{11.17} & 6239 & 24.35 & 16.90 & 12.89 \\
    \bottomrule
  \end{tabular}
\end{table}

\section{Calibration Under Compression (ECE)}\label{app:ece}
\begin{figure}[htbp]
  \centering
  \includegraphics[width=0.62\linewidth]{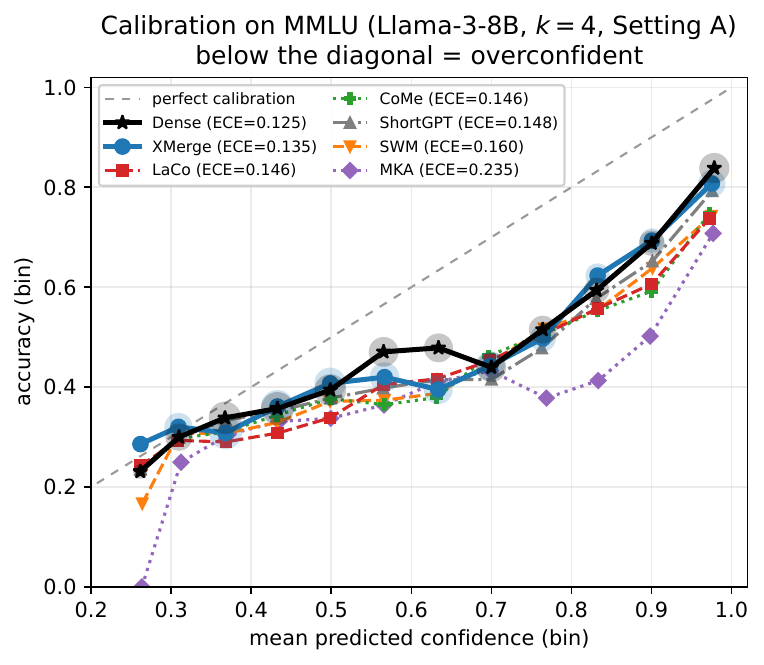}
  \caption{Reliability diagrams on MMLU (Llama-3-8B, $k{=}4$, Setting~A) for all seven methods: per-bin accuracy
  vs.\ mean confidence; the diagonal is perfect calibration and points below it are overconfident. \XMerge
  (ECE $0.135$) tracks the dense model ($0.125$) most closely, the two highlighted in bold with
  bin-population shading; the analytic folds deviate further, LaCo and CoMe ($0.146$), ShortGPT ($0.148$)
  and SWM ($0.160$), and MKA ($0.235$) collapses below the diagonal in the high-confidence bins. Marker
  area $\propto$ bin population.}
  \label{fig:ece}
\end{figure}

As a first probe of trustworthiness beyond task accuracy (\S\ref{sec:limits}), we measure expected calibration
error (ECE) on Llama-3-8B at $k{=}4$ under the recovery-free Setting~A protocol. ECE is a per-model property
(the gap between a model's stated confidence and its actual accuracy), orthogonal to accuracy itself. We reuse
the existing multiple-choice pipeline (\S\ref{app:eval}): each MMLU choice is scored by mean per-token NLL, the
length-normalized posterior is $\mathrm{softmax}(-\overline{\mathrm{nll}})$, confidence is its maximum, and ECE
bins confidence into 15 equal-width bins on $[0,1]$ (all $14{,}042$ MMLU items; the 12 with a degenerate empty
choice span are excluded identically for every method). Table~\ref{tab:ece} reports accuracy, ECE, MCE (maximum calibration error), signed
over-confidence (confidence minus accuracy; positive $=$ overconfident), and $\Delta\mathrm{ECE}$ relative to
the dense model; Fig.~\ref{fig:ece} shows the reliability diagrams.

\XMerge has the lowest ECE degradation of all six operators ($\Delta\mathrm{ECE}{=}{+}0.010$), roughly half
that of the next-best operator and an order of magnitude below the analytic similarity merge MKA
($+0.110$, also the most miscalibrated by MCE). Every method is overconfident, so compression does not improve
calibration in absolute terms; the finding is that \emph{reconstructive} merging preserves the dense model's
calibration best. As a single backbone and metric this result does not establish a general safety guarantee
(\S\ref{sec:limits}).

\begin{table}[htbp]
  \centering\small
  \caption{Calibration on MMLU (Llama-3-8B, $k{=}4$, Setting~A). ECE/MCE lower is better; signed gap $>0$ $=$
  overconfident; $\Delta$ECE is relative to the dense model. \XMerge is the best-calibrated operator.}
  \label{tab:ece}
  \begin{tabular}{lccccc}
    \toprule
    Method & Acc & ECE $\downarrow$ & MCE $\downarrow$ & Signed gap & $\Delta$ECE vs dense \\
    \midrule
    Dense            & 0.467 & 0.125 & 0.260 & $+0.125$ & --- \\
    ShortGPT         & 0.427 & 0.148 & 0.285 & $+0.148$ & $+0.023$ \\
    LaCo             & 0.376 & 0.146 & 0.294 & $+0.146$ & $+0.021$ \\
    MKA              & 0.411 & 0.235 & 0.421 & $+0.235$ & $+0.110$ \\
    SWM              & 0.410 & 0.160 & 0.279 & $+0.160$ & $+0.035$ \\
    CoMe             & 0.392 & 0.146 & 0.308 & $+0.146$ & $+0.021$ \\
    \textbf{\XMerge} & 0.434 & \textbf{0.135} & \textbf{0.267} & $+0.134$ & \textbf{+0.010} \\
    \bottomrule
  \end{tabular}
\end{table}

\section{Baseline Implementation Audit}\label{app:audit}
All five baselines are official implementations or faithfully audited ports
(\texttt{baselines\_\{shortgpt,laco,mka,swm,come\}.py}). Each uses its authors' native selection and
structural operator at the same target depth; the only shared inputs are the model and the WikiText-2 train
calibration corpus (used by baselines for selection/merge ratios only, no weight training). Each baseline
follows the algorithm in its cited publication.

\paragraph{ShortGPT.} BI-guided (angular block-influence) layer drop; verified faithful against the official
implementation (matching selected layers). No deviation was identified.

\paragraph{LaCo.} Bottom-up layer collapse folding parameter deltas into a surviving block. Minor
compatibility nits only. \emph{Note (not independent from SWM at $k{=}1$):} on Llama-3-8B $k{=}1$, LaCo and
SWM produce a \emph{byte-identical} model: both select the pair $[10,11]$ and both apply the same
$\text{base}\mathrel{+}=(\text{src}-\text{base})$ fold (SWM's \texttt{weight\_factor}$=1$ equals LaCo's
single fold). They diverge at $k{=}2/4$ as SWM's window selects different spans. We disclose this so the two
are not read as independent points at $k{=}1$.

\paragraph{MKA.} Top-down analytical merge of the most-similar deepest adjacent layers. Our initial port
selected by argmax-NPIB, which merged near-\emph{input} layers and produced an artificial collapse; we
corrected it to the official top-down rule (layers $N{-}2,N{-}1$), and all reported MKA results use this
corrected port. The corrected port's high PPL at depth (e.g.\ $8343$ on Llama-3.2-1B
$k{=}4$) is genuine top-down behavior on small models, reproduced from the official rule.

\paragraph{SWM.} Sliding-window merge by weighted parameter averaging. At 8B the second (reference) model
copy was CPU-offloaded to fit 40\,GB MIG memory; results are numerically identical to the on-GPU path. In
Setting~B, SWM is healed under the same uniform KD as every other healed method (SWM's native LoRA recovery is reported
separately as a disclosed, non-comparable supplementary, \S\ref{app:recovery}).

\paragraph{CoMe.} Merge by attention head-group concatenation; its native hierarchical distillation is
deferred to Setting~B (where CoMe receives the same recovery budget as all methods). On Qwen2.5-0.5B (2 KV
heads) head-group concatenation degenerates: \texttt{distribute\_and\_round(2,\,ratio)} yields groups
$[2,0]$, so each merge takes attention entirely from the dominant layer, driving the $k{=}4$ collapse.
On the 8-KV-head models this degeneracy does not occur, yet CoMe still collapses on CORE at $k{=}4$ (e.g.\
Qwen3-8B $0.171$), confirming a real operator weakness rather than a head-count artifact.

\section{Selector Definitions and Layer Choices}\label{app:selector}
The deployed selector removes the eligible block with the smallest $s_l=\max(\zRM_l,\zBI_l)$, greedily,
with selection statistics fixed on the pristine model (\S\ref{sec:iterative}). Table~\ref{tab:selfimit}
lists the layers removed at $k{=}4$ by RM alone vs.\ max-fusion; max-fusion is byte-identical to
RM on 5/7 backbones and changes selection only on Qwen3-0.6B ($k{=}4$) and Qwen3-1.7B, precisely where the
two axes decorrelate: $\mathrm{corr}(\RM,\BI)\approx0.01$ on Qwen3-1.7B (where it acts) vs.\ $\approx0.74$
on Qwen3-8B (where it is a no-op).

\paragraph{Merge direction (which neighbor survives).} Given the selected block $f_\ell$, \XMerge merges it
with the left or right neighbor and keeps the neighbor as the surviving block. When both neighbors are
eligible, we pick the side maximizing an adjacent \emph{activation-patching redundancy} score
$S_{\mathrm{patch}}$: for each calibration example we corrupt the input, restore each layer's attention and
MLP outputs to their clean values one layer at a time, and record the per-layer logit recovery;
$S_{\mathrm{patch}}[l]$ is the Pearson correlation between the recovery profiles of adjacent layers $l$ and
$l{+}1$ (high correlation = the two blocks play redundant causal roles, so absorbing one into the other
preserves function). This score sets only the merge \emph{direction}; the block removed is fixed by the
cross-axis selector above.

\paragraph{Why RM and BI (7-criterion oracle backtest).} Table~\ref{tab:oracle} reports the rank correlation
($|\rho|$, Spearman) between each of seven candidate criteria and an oracle layer-removal sweep, on two
backbones (Llama-3.2-1B, Qwen2.5-0.5B). RM is the strongest CORE predictor on
\emph{both} models, and BI is close behind on both (second on Qwen). No other criterion is robust across both
architectures: WS rivals RM on Llama ($0.834$) but collapses on Qwen ($0.212$), the mirror image of
Taylor's architecture-specific behaviour. The remaining criteria are strong \emph{perplexity} predictors but
weak CORE predictors: P (a causal layer-importance / patching-recovery score from prior
work~\cite{quantdamage2026}), PMR, and Taylor all reach $|\rho|\!=\!0.6$--$0.8$ against PPL yet
$|\rho|\!\le\!0.33$ against CORE. Causal and first-order importance identify which layers a computation
\emph{depends on} and track LM loss, but do not predict which adjacent pair is safe to \emph{merge} for
retained downstream ability, a negative transfer that motivates the magnitude/angle signals we use; IRS is
uninformative on both. RM and BI are near-tied as CORE predictors, so we adopt their fusion for robustness,
not an ``RM$>$BI'' claim.

\begin{table}[htbp]
  \centering\small
  \caption{Oracle backtest: absolute Spearman rank correlation $|\rho|$ between each candidate selection
  criterion and the oracle removal sweep, on two backbones. Higher = better predictor of that metric. RM is
  the top CORE predictor on both models and BI is close behind on both; WS is strong on Llama but not Qwen;
  P/PMR/Taylor predict perplexity but not CORE; IRS is uninformative. Bold = best CORE predictor per model.}
  \label{tab:oracle}
  \begin{tabular}{lcccc}
    \toprule
    & \multicolumn{2}{c}{Llama-3.2-1B} & \multicolumn{2}{c}{Qwen2.5-0.5B} \\
    \cmidrule(lr){2-3}\cmidrule(lr){4-5}
    Criterion & $|\rho|$ CORE & $|\rho|$ PPL & $|\rho|$ CORE & $|\rho|$ PPL \\
    \midrule
    RM (magnitude) & \textbf{0.849} & 0.107 & \textbf{0.325} & 0.806 \\
    BI (angular)   & 0.805 & 0.043 & 0.321 & 0.508 \\
    WS             & 0.834 & 0.224 & 0.212 & 0.138 \\
    P (causal patch)~\cite{quantdamage2026} & 0.328 & 0.628 & 0.057 & 0.370 \\
    PMR            & 0.314 & 0.810 & 0.232 & 0.793 \\
    Taylor         & 0.283 & 0.589 & 0.221 & 0.736 \\
    IRS            & 0.121 & 0.024 & 0.012 & 0.262 \\
    \bottomrule
  \end{tabular}
\end{table}

\begin{table}[htbp]
  \centering\small
  \caption{Layers removed at $k{=}4$ (0-indexed), RM-only vs.\ deployed max-fusion. ``chg'' marks the two
  models where max-fusion overrides RM.}
  \label{tab:selfimit}
  \begin{tabular}{llll}
    \toprule
    Model (L) & RM removes & max-fusion removes & \\
    \midrule
    Llama-3.2-1B (16) & \{11,12,13,14\} & \{11,12,13,14\} & \\
    Llama-3.2-3B (28) & \{21,22,23,24\} & \{21,22,23,24\} & \\
    Llama-3-8B (32)   & \{23,24,25,26\} & \{23,24,25,26\} & \\
    Qwen2.5-0.5B (24) & \{9,10,12,13\}  & \{9,10,12,13\}  & \\
    Qwen3-0.6B (28)   & \{3,12,13,14\}  & \{12,13,14,15\}  & chg \\
    Qwen3-1.7B (28)   & \{3,4,5,7\}     & \{11,12,13,14\}  & chg \\
    Qwen3-8B (36)     & \{18,31,32,33\} & \{18,31,32,33\} & \\
    \bottomrule
  \end{tabular}
\end{table}

\section{Recovery Protocol}\label{app:recovery}
\paragraph{Uniform healer (Setting B).} Every method's $k{=}4$ compressed cell is healed by the \emph{same}
procedure: LoRA adapters~\cite{hu2022lora} (rank 16, $\alpha$ 32) on a forward
knowledge-distillation~\cite{hinton2015distillation} objective
$D_{\mathrm{KL}}(\text{teacher}\,\|\,\text{student})$ over next-token logits against the original FP16 model
as teacher, on WikiText-2 \emph{train}, in bf16 with 50 warmup steps. After training, the LoRA is merged
back, so the healed model adds \emph{zero} inference-time parameters. The compute budget is
\emph{iso-compute}: each method is healed for the same wall-clock as \XMerge's own $k{=}4$ compression cost
on that model (Llama-3.2-1B $4968$\,s; Qwen3-8B $14039$\,s). Per-model recipe: 1B uses batch size 4,
$4096$ calibration sequences, no gradient checkpointing; 8B uses batch size 2 with gradient checkpointing.
Each method is run at all three LRs in the shared grid $\{1,2,5\}\times10^{-5}$, and the main-text table
reports the \emph{mean} over the three LRs rather than selecting the best rate
on the evaluation metric; per-LR values are in Table~\ref{tab:recovery-perlr}. Because the budget is
wall-clock-capped, the step count is set by the budget rather than fixed: a representative healed cell runs
$\approx$15{,}600 steps ($\approx$16 epochs over $4096\times512$-token sequences, $\approx$32M training
tokens) on Llama-3.2-1B, and $\approx$12{,}800 steps ($\approx$7 epochs, $\approx$13M tokens) on Qwen3-8B.
The only trainable parameters are the rank-16 LoRA adapters, which are merged into the weights afterward
(zero added parameters at inference).

\begin{table}[htbp]
  \centering\small
  \caption{Per-LR recovery (Setting~B, $k{=}4$); Table~\ref{tab:recovery} reports the per-method mean of these
  three rows. CORE and MMLU to three decimals, wiki-PPL to two.}
  \label{tab:recovery-perlr}
  \begin{tabular}{llccc}
    \toprule
    Method & LR & CORE & wiki-PPL & MMLU \\
    \midrule
    \multicolumn{5}{l}{\emph{Llama-3.2-1B}} \\
    \XMerge   & $1{\times}10^{-5}$ & .255 & 15.54 & .300 \\
    \XMerge   & $2{\times}10^{-5}$ & .255 & 15.47 & .298 \\
    \XMerge   & $5{\times}10^{-5}$ & .247 & 15.28 & .296 \\
    ShortGPT  & $1{\times}10^{-5}$ & .250 & 16.99 & .294 \\
    ShortGPT  & $2{\times}10^{-5}$ & .246 & 16.51 & .297 \\
    ShortGPT  & $5{\times}10^{-5}$ & .240 & 16.18 & .294 \\
    CoMe      & $1{\times}10^{-5}$ & .161 & 17.38 & .285 \\
    CoMe      & $2{\times}10^{-5}$ & .172 & 16.96 & .286 \\
    CoMe      & $5{\times}10^{-5}$ & .176 & 16.98 & .286 \\
    SWM       & $1{\times}10^{-5}$ & .225 & 15.55 & .296 \\
    SWM       & $2{\times}10^{-5}$ & .227 & 15.23 & .296 \\
    SWM       & $5{\times}10^{-5}$ & .216 & 15.11 & .296 \\
    \midrule
    \multicolumn{5}{l}{\emph{Qwen3-8B}} \\
    \XMerge   & $1{\times}10^{-5}$ & .351 & 10.90 & .327 \\
    \XMerge   & $2{\times}10^{-5}$ & .363 & 10.75 & .327 \\
    \XMerge   & $5{\times}10^{-5}$ & .359 & 10.33 & .326 \\
    ShortGPT  & $1{\times}10^{-5}$ & .347 & 11.96 & .320 \\
    ShortGPT  & $2{\times}10^{-5}$ & .350 & 11.55 & .324 \\
    ShortGPT  & $5{\times}10^{-5}$ & .356 & 11.19 & .318 \\
    CoMe      & $1{\times}10^{-5}$ & .344 & 9.21 & .333 \\
    CoMe      & $2{\times}10^{-5}$ & .347 & 9.19 & .332 \\
    CoMe      & $5{\times}10^{-5}$ & .337 & 9.30 & .328 \\
    SWM       & $1{\times}10^{-5}$ & .365 & 12.35 & .319 \\
    SWM       & $2{\times}10^{-5}$ & .350 & 12.05 & .316 \\
    SWM       & $5{\times}10^{-5}$ & .359 & 11.27 & .318 \\
    \bottomrule
  \end{tabular}
\end{table}

\paragraph{SWM-native disclosure (non-comparable).} SWM ships a native recovery that is
\emph{instruction tuning} (Alpaca, LoRA, cross-entropy), not the uniform WikiText-2 KD used above. Under it,
the SWM cell heals to CORE \emph{above the dense model} (Qwen3-8B up to $0.490$ vs.\ dense $0.417$), a
broad per-task lift (largest on boolq/winograd/copa/commonsense\_qa, monotonic in LR) characteristic of
added instruction-following supervision, not merge recovery. We therefore exclude SWM-native from the fair
4-way comparison and report it only as a disclosed supplementary; the comparable SWM row uses the same
uniform KD as all methods.

\section{Task-Regime Breakdown}\label{app:regime}
CORE's 22 tasks split by \texttt{num\_fewshot} into a 0-shot bucket (7 tasks: copa, coqa,
hellaswag\_zeroshot, lambada\_openai, openbook\_qa, winograd, winogrande) and a few-shot/ICL bucket (15
tasks: arc\_challenge, arc\_easy, boolq, commonsense\_qa, hellaswag, piqa, squad, jeopardy, agi\_eval\_lsat\_ar
[3-shot], and 6 BIG-bench tasks [10-shot]). Values are centered CORE (per-task scores that average to
\texttt{core\_metric}), $k{=}4$. We compare methods \emph{within} a regime (task set fixed); the two buckets
have different task counts, so cross-bucket gaps are not comparable. A \emph{collapse} is a cell with
centered CORE $<0.10$.

\begin{table}[htbp]
  \centering\small
  \caption{0-shot regime (7 tasks), centered CORE, $k{=}4$. Bold = best in row.}
  \begin{tabular}{lcccccc}
    \toprule
    Model & ShortGPT & LaCo & MKA & SWM & CoMe & XM \\
    \midrule
    Llama-3.2-1B & .140 & .071 & .020 & .139 & .038 & \textbf{.236} \\
    Llama-3.2-3B & .421 & .251 & .350 & .398 & .297 & \textbf{.450} \\
    Llama-3-8B   & .489 & .435 & .374 & .444 & .403 & \textbf{.493} \\
    Qwen2.5-0.5B & .170 & .149 & .134 & .148 & .117 & \textbf{.192} \\
    Qwen3-0.6B   & .071 & .135 & .106 & .093 & .044 & \textbf{.177} \\
    Qwen3-1.7B   & .156 & .182 & .196 & .188 & .020 & \textbf{.251} \\
    Qwen3-8B     & .205 & .210 & .167 & .188 & .209 & \textbf{.237} \\
    \midrule
    Mean         & .236 & .205 & .192 & .228 & .161 & \textbf{.291} \\
    \bottomrule
  \end{tabular}
\end{table}

\begin{table}[htbp]
  \centering\small
  \caption{Few-shot / ICL regime (15 tasks), centered CORE, $k{=}4$. Bold = best in row.}
  \begin{tabular}{lcccccc}
    \toprule
    Model & ShortGPT & LaCo & MKA & SWM & CoMe & XM \\
    \midrule
    Llama-3.2-1B & .097 & .064 & $-$.022 & .124 & $-$.025 & \textbf{.184} \\
    Llama-3.2-3B & .350 & .194 & .321 & .266 & .271 & \textbf{.385} \\
    Llama-3-8B   & .546 & .322 & .392 & .358 & .484 & \textbf{.553} \\
    Qwen2.5-0.5B & .167 & .162 & .147 & .139 & .085 & \textbf{.172} \\
    Qwen3-0.6B   & .083 & .172 & \textbf{.193} & .120 & .003 & .179 \\
    Qwen3-1.7B   & .171 & .207 & \textbf{.333} & .255 & .006 & .281 \\
    Qwen3-8B     & .281 & .259 & .280 & .296 & .154 & \textbf{.344} \\
    \midrule
    Mean         & .242 & .197 & .235 & .222 & .140 & \textbf{.300} \\
    \bottomrule
  \end{tabular}
\end{table}

\section{Measured Inference Speedup}\label{app:latency}
At a fixed removal count $k$ every depth-compression operator (ours and all baselines) yields the
\emph{same} standard $L{-}k$-block decoder with zero added inference parameters (\S\ref{app:cost}), so
inference latency is not a per-method quantity: it is a single per-$(\text{backbone},k)$ value. We therefore
benchmark it once per $(\text{backbone},k)$ on the real \XMerge checkpoint ($k{=}0$ is dense) and it stands
for all methods. Table~\ref{tab:latency} reports \emph{measured} decode speedup versus dense, alongside the
depth-proportional prediction $L/(L{-}k)$.

\paragraph{Protocol.} NVIDIA A100-SXM4-80GB, single \texttt{MIG 3g.40gb} slice (3/7 of the SMs, not a full
A100), driver 550.127.08, CUDA 12.8, PyTorch 2.10.0, Transformers 5.11.0, fp16. HuggingFace eager
\texttt{generate}, greedy decoding, prompt length 512 and exactly 256 forced new tokens; decode throughput is
$\text{batch}\cdot\text{gen\_len}/(t_{\text{total}}-t_{\text{prefill}})$. \textbf{Batch~1 (interactive
latency) is primary}: 20 timed reps after 2 warmups, median reported, run-to-run CV $\le 4\%$ on all seven
backbones. \textbf{Batch~16 (throughput) is secondary}, reported as median speedup only: a MIG slice
partitions SMs/L2/bandwidth but shares the physical card's power/thermal domain with co-tenants we cannot see,
so sustained batched runs on the large models downclock and carry high CV ($\sim$25--30\%); on a dedicated
A100 this variance would be far lower. Three caveats stated up front: (i)~latency at fixed $k$ is
method-independent (shared architecture); (ii)~MIG does not isolate power/thermal, hence the batch-16
variance; (iii)~the depth-proportional \emph{ratio} ($\approx L/(L{-}k)$) is more portable across serving systems than the
absolute tok/s, which under HF eager on a MIG slice understates a production engine (vLLM/TensorRT-LLM).

\begin{table}[htbp]
  \centering\small
  \caption{Measured decode speedup vs.\ dense, $k{=}1/2/4$ per cell. ``Dense'' is batch-1 decode tok/s on the
  MIG slice; ``Theory'' is $L/(L{-}k)$. Measured batch-1 tracks the depth-proportional prediction; batch-16
  is the high-variance secondary regime (\S\ref{app:latency}).}
  \label{tab:latency}
  \begin{tabular}{lccccc}
    \toprule
    Model & $L$ & Dense (tok/s) & Theory $k{=}1/2/4$ & Batch-1 $k{=}1/2/4$ & Batch-16 $k{=}1/2/4$ \\
    \midrule
    Llama-3.2-1B & 16 & 89.6 & 1.07/1.14/1.33 & 1.05/1.12/1.21 & 1.04/1.11/1.27 \\
    Llama-3.2-3B & 28 & 50.9 & 1.04/1.08/1.17 & 1.05/1.11/1.17 & 1.07/1.10/1.18 \\
    Llama-3-8B   & 32 & 41.3 & 1.03/1.07/1.14 & 1.03/1.06/1.12 & 1.03/1.06/1.13 \\
    Qwen2.5-0.5B & 24 & 59.1 & 1.04/1.09/1.20 & 1.04/1.10/1.21 & 1.02/1.07/1.17 \\
    Qwen3-0.6B   & 28 & 42.7 & 1.04/1.08/1.17 & 1.02/1.06/1.14 & 1.04/1.09/1.17 \\
    Qwen3-1.7B   & 28 & 40.1 & 1.04/1.08/1.17 & 1.10/1.15/1.23 & 1.08/1.03/1.11 \\
    Qwen3-8B     & 36 & 30.4 & 1.03/1.06/1.12 & 1.02/1.06/1.13 & 1.03/1.06/1.11 \\
    \midrule
    \textbf{Mean} & --- & --- & --- & \textbf{1.04/1.09/1.17} & \textbf{1.04/1.07/1.16} \\
    \bottomrule
  \end{tabular}
\end{table}

Measured batch-1 speedup tracks $L/(L{-}k)$ closely (mean $1.04/1.09/1.17\times$ at $k{=}1/2/4$; the largest
per-cell deviations are on the shallowest backbones, where a single removed layer is a larger fraction of
depth and prefill/sampling overheads are relatively larger). Because the speedup is depth-proportional and
method-independent, the quality differences between operators at matched $k$ are obtained at \emph{identical}
inference cost, so the quality--latency Pareto frontier is determined entirely by which operator retains
the most quality per removed layer, which is exactly the per-backbone comparison plotted in the main text
(Fig.~\ref{fig:pareto}).

\section{Construction-Time Measurements}\label{app:cost}
One-time compression cost in seconds, reported as $k{=}1/2/4$ per cell. Training-free drop/merge baselines
are seconds--minutes; \XMerge is minutes--hours, scaling $\approx$linearly with $k$ ($\approx$300
boundary-optimization steps per merge). \XMerge adds \emph{zero} inference-time parameters; at matched depth
all methods share the same $N{-}k$-layer shape, so parameter count, FLOPs/token, and memory are identical
across methods and are a single per-$k$ quantity (not per method). We additionally \emph{measure} decode
throughput directly (\S\ref{app:latency}): the resulting speedup is depth-proportional ($\approx L/(L{-}k)$,
mean $1.04/1.09/1.17\times$ at $k{=}1/2/4$, batch~1) and, like the parameter/FLOP counts, is a single
per-$(\text{backbone},k)$ quantity shared by all operators. Construction timings below are wall-clock on a
single NVIDIA A100-SXM4-80GB (MIG slice).

\begin{table}[htbp]
  \centering\small
  \caption{Construction cost (s), $k{=}1/2/4$ per cell. A few training-free-baseline cells are non-monotonic
  in $k$ (e.g.\ LaCo and CoMe on Llama-3-8B, where the $k{=}1$ entry exceeds $k{=}2$): these are one-off
  first-call effects (weight load, CPU-offload paging, and kernel/cache warm-up on the largest backbones), not
  algorithmic scaling, and do not affect any reported quality number.}
  \resizebox{\linewidth}{!}{%
  \begin{tabular}{lcccccc}
    \toprule
    Model & ShortGPT & LaCo & MKA & SWM & CoMe & XM \\
    \midrule
    Llama-3.2-1B & 22/21/20   & 31/36/250   & 28/26/25  & 70/171/370   & 44/67/87    & 1261/2456/4968 \\
    Llama-3.2-3B & 36/48/35   & 70/89/109   & 51/51/44  & 140/380/544  & 85/116/162  & 1978/3800/7509 \\
    Llama-3-8B   & 119/159/111& 1172/158/214& 96/100/99 & 216/846/1149 & 1196/224/329& 3975/7832/15783 \\
    Qwen2.5-0.5B & 14/12/13   & 34/46/70    & 25/24/19  & 49/65/207    & 34/50/57    & 375/712/1402 \\
    Qwen3-0.6B   & 15/17/16   & 42/62/109   & 23/24/26  & 96/269/417   & 44/52/67    & 424/806/1553 \\
    Qwen3-1.7B   & 25/27/25   & 59/64/130   & 32/37/33  & 96/223/321   & 59/81/118   & 1026/1998/4106 \\
    Qwen3-8B     & 84/147/96  & 143/223/261 & 120/109/177& 222/300/621 & 200/583/562 & 3587/7121/14039 \\
    \bottomrule
  \end{tabular}%
  }
\end{table}

\paragraph{One-time cost vs.\ recurring saving: break-even.} The construction cost above is paid
\emph{once}; the decode speedup of \S\ref{app:latency} is paid back on \emph{every} generated token for the
life of the deployment. Relative to the most expensive training-free baseline (SWM) at $k{=}4$, \XMerge's
one-time cost is $3.7$--$22.6\times$ larger in wall-clock, but in absolute terms it is only minutes to $4.4$
hours (Table above), and it amortizes quickly. Let $T_c$ be the $k{=}4$ construction time, $s_d$ the dense
batch-1 decode rate (Table~\ref{tab:latency}), and $\sigma$ the measured batch-1 speedup; the per-token
wall-clock saving is $s_d^{-1}(1-\sigma^{-1})$, so \XMerge pays for itself after
$N^\star = T_c\, s_d\, \sigma/(\sigma-1)$ generated tokens (Table~\ref{tab:breakeven}). Across the seven
backbones $N^\star$ ranges from $0.48$M to $6.1$M generated tokens (equivalently $\approx 1.9$k--$24$k
requests of 256 tokens each), after which every further token is net faster than dense. This is a small
fraction of any production serving lifetime, so the one-time cost is recovered almost immediately and the
speedup dominates thereafter. Because $N^\star$ depends on $T_c$ and $s_d$ measured on the \emph{same} MIG
slice, it is to first order invariant to the absolute speed of the serving stack (a faster engine shrinks
$T_c$ and grows $s_d$ together); a production engine that accelerates decode more than the gradient-descent
construction would raise $N^\star$ somewhat, but even a $3\times$ understatement keeps break-even in the low
tens of thousands of requests.

\begin{table}[htbp]
  \centering\small
  \caption{Break-even for the one-time construction cost at $k{=}4$. $T_c$ = construction time (s, from the
  table above); dense $s_d$ and speedup $\sigma$ are the batch-1 values from Table~\ref{tab:latency};
  $N^\star = T_c\, s_d\, \sigma/(\sigma-1)$ is the number of \emph{generated} tokens after which the recurring
  decode saving offsets the construction cost, also shown as 256-token requests. After $N^\star$, every further token is
  net faster than dense.}
  \label{tab:breakeven}
  \begin{tabular}{lccccc}
    \toprule
    Model & $T_c$ (s) & Dense $s_d$ (tok/s) & $\sigma$ ($k{=}4$) & $N^\star$ (M tok) & $N^\star$ (k req) \\
    \midrule
    Llama-3.2-1B & 4968  & 89.6 & 1.21 & 2.56 & 10.0 \\
    Llama-3.2-3B & 7509  & 50.9 & 1.17 & 2.63 & 10.3 \\
    Llama-3-8B   & 15783 & 41.3 & 1.12 & 6.08 & 23.8 \\
    Qwen2.5-0.5B & 1402  & 59.1 & 1.21 & 0.48 & 1.9 \\
    Qwen3-0.6B   & 1553  & 42.7 & 1.14 & 0.54 & 2.1 \\
    Qwen3-1.7B   & 4106  & 40.1 & 1.23 & 0.88 & 3.4 \\
    Qwen3-8B     & 14039 & 30.4 & 1.13 & 3.71 & 14.5 \\
    \bottomrule
  \end{tabular}
\end{table}

\end{document}